\documentclass[11pt]{article}

\usepackage[preprint]{acl}

\usepackage{times}
\usepackage{latexsym}

\usepackage[T1]{fontenc}

\usepackage[utf8]{inputenc}
\usepackage{array}
\usepackage{tabularx}
\usepackage{booktabs}
\usepackage{multirow}
\usepackage{todonotes}
\usepackage{xurl} 
\usepackage{microtype}

\usepackage{inconsolata}

\usepackage{graphicx}

\title{Why Do LLMs Accept Generics? \\ Explaining Acceptability using Psycholinguistic and Model-Derived Factors} 

\author{
  \textbf{Mervat Abassy\textsuperscript{1}} 
  \textbf{Zhaoxin Feng\textsuperscript{2}} 
  \textbf{Tatsuki Kuribayashi\textsuperscript{1,3}} 
  \textbf{Kentaro Inui\textsuperscript{1,3,4}} 
  \textbf{Emmanuele Chersoni\textsuperscript{2}}
\\
\\
  \textsuperscript{1}MBZUAI  
  \textsuperscript{2}The Hong Kong Polytechnic University
  \\
  \textsuperscript{3}Tohoku University
  \textsuperscript{4}RIKEN
\\
  \texttt{\{mervat.abassy\}@mbzuai.ac.ae}
}

\begin{document}
\maketitle
\begin{abstract}
Psycholinguistic studies explore why humans accept some generic statements but reject others by showing that these judgments depend on more than how common a property is. People consider several other factors such as how well it sets the kind apart from others or whether it is characteristic of the kind. 

Inspired by the linguistic and cognitive literature, we explore to what extent factors from these studies explain how large language models (LLMs) judge generics. Our results with twelve models on a psycholinguistic dataset and on naturally occurring generics show that models weigh how common a property is against how well it distinguishes the kind, and combine the two much as people do, though less clearly. Like people, they favor characteristic properties, but unlike the accounts of human judgment, they ignore striking ones. What matters most to LLMs is whether a statement is factually true, and much of the influence of human factors reaches their ratings through that judgment. In six 7--9B models, shifting the hidden states along each factor's direction moves the judgment as the factor predicts.\footnote{Code and artifacts will be released.} 

\end{abstract}

\section{Introduction}

People readily accept ``Lions have manes'' but not ``Lions are male,'' even though the latter is true of more lions than the former. This contrast suggests that judgments about generic statements, which concern types rather than individuals, cannot be explained simply by the frequency of a property or the number of individuals for which a statement is true~\citep{leslie-2008-generics,prasada-etal-2013-conceptual}. Instead, these judgments reflect how people represent categories. Generics have been argued to express the mind's default mode of generalization~\citep{leslie-2008-generics}. People accept them on little evidence, particularly when the property is dangerous~\citep{cimpian-etal-2010-generic}, treat properties that are principled consequences of being a kind differently from merely common ones~\citep{prasada-etal-2013-conceptual}, and even extend characteristic generics to universal claims such as ``All ducks lay eggs''~\citep{leslie-etal-2011-ducks}. \citet{ryu-etal-2022-flexible} showed that the acceptability of generics depends on two interacting factors: \textbf{prevalence}, or how common a property is within a kind, and \textbf{cue validity}, or how strongly that property distinguishes the kind from relevant alternatives. 
Other accounts of how people judge generics (e.g. \citet{hoorens-etal-2026-constat,hermans-etal-2026-justifiable}) treat people as intuitive statisticians who weigh how common a property is in the kind against how common it is in alternative
kinds: a highly diagnostic property can support a generic statement even when it is relatively uncommon within the kind. Having a mane distinguishes lions from other animals, whereas being male does not.

\begin{figure}[t]
\centering
\includegraphics[width=\columnwidth]{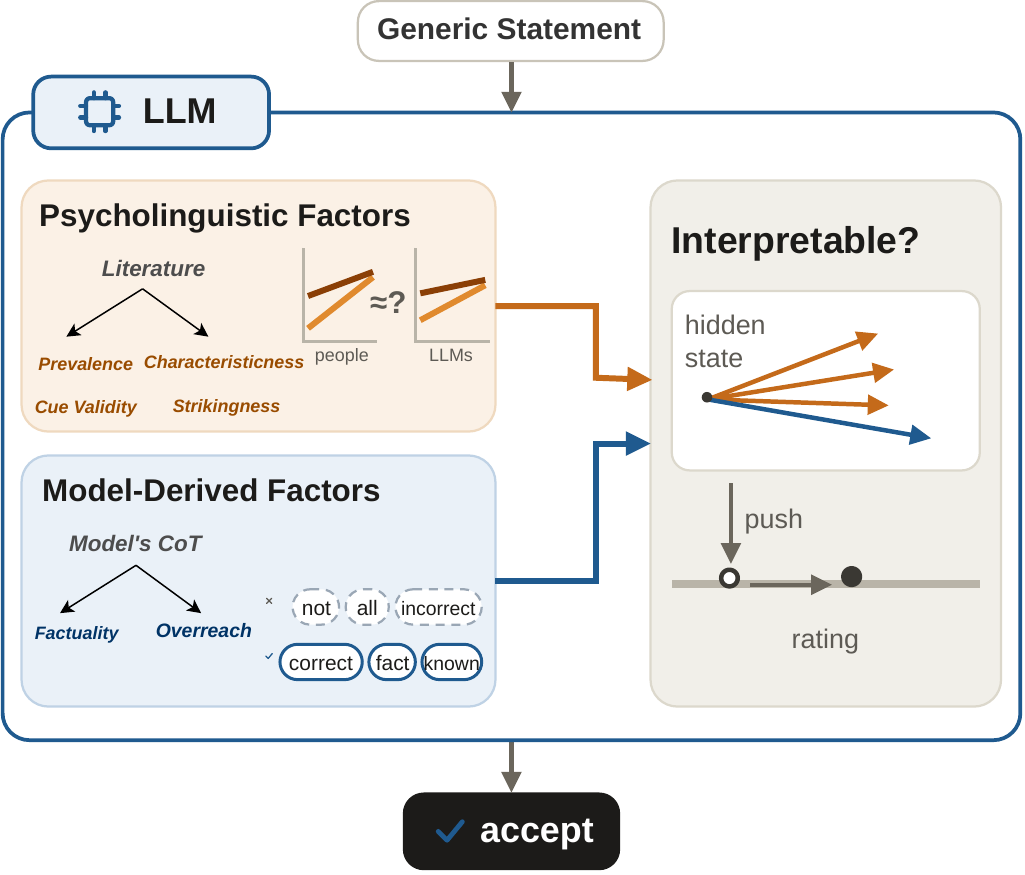}
\caption{\label{fig:overview}
  We explore how far factors from the psycholinguistic literature explain LLM judgments of generics, which further factors the models point to, and whether these factors can steer acceptability inside the models.}
\end{figure}

Large language models (LLMs) are increasingly used as proxies for human judgments and as models of human language processing~\citep{binz-schulz-2023-using,dillion-etal-2023-can}. Their agreement with human judgments, however, does not establish that models arrive at those judgments in the same way as people. Generics have long posed a challenge for theories of language acquisition: although their truth conditions are difficult to specify, children understand them at an early age and with little explicit evidence~\citep{leslie-2008-generics}. LLMs offer a useful comparison because they learn from text, in which generics are common. 
Previous work on generics in LLMs has examined whether models overgeneralize or accept the same statements as humans~\citep{ralethe-buys-2022-generic}, as well as how generics relate to quantified statements~\citep{cilleruelo-etal-2025-generics}. It remains unclear, however, whether models \textit{rely on the same factors that shape human judgments}. We thus ask the following questions:
\begin{itemize}
    \item To what extent can LLM acceptability be explained by factors used in human psycholinguistic studies;
  \item What additional factors the models themselves point to when explaining their judgments, and how they relate to the psychological ones;
  \item Whether shifting the models’ internal states along each factor’s direction changes their judgments as the factor predicts.
\end{itemize}
\section{Related work}
\label{sec:related}

What makes generic statements acceptable has long been a central question for linguists. \citet{leslie-2008-generics} argued that people's acceptance of generics depends on more than prevalence, or how common a property is within a kind. \citet{ryu-etal-2022-flexible} further showed that acceptance also depends on cue validity, or how strongly a property predicts kind membership, while \citet{hoorens-etal-2026-constat} and \citet{hermans-etal-2026-justifiable} highlighted the role of relative prevalence, showing that acceptance depends on how common a property is in one kind compared with others. However, people's judgments reflect more than statistical patterns. Characteristic properties are judged differently from merely common ones \citep{prasada-etal-2013-conceptual}, sometimes supporting universal claims such as \emph{All ducks lay eggs} \citep{leslie-etal-2011-ducks}. Similarly, \citet{leslie-2008-generics} argued that striking properties require little evidence to support a generic, and \citet{cimpian-etal-2010-generic} showed that dangerous properties are accepted at lower prevalence levels, although this effect seems to be limited to intermediate prevalence \citep{hermans-etal-2026-justifiable}.

Research on language models has also engaged with generics. \citet{ralethe-buys-2022-generic} and \citet{allaway-etal-2024-exceptions} found that models overgeneralize from generics to universal claims and struggle with exceptions. Although models show human-like nonmonotonic reasoning when given exceptions, their judgments often shift when unrelated or even supporting examples are added \citep{leidinger-etal-2024-llms}. In how they judge generics, \citet{peters-etal-2026-science} found that LLMs rate scientific generics as more generalizable and credible than people and \citet{hu-etal-2024-failures} showed that they largely fail to distinguish principled from statistical properties once prevalence is controlled. These studies 
leave open the issue of which factors shape models' judgments and whether they resemble those underlying human judgments.

A second line of work infers model judgments from probabilities rather than written answers. Direct probability measures can reveal knowledge that metalinguistic prompts miss \citep{hu-levy-2023-prompting,kauf-etal-2024-log} and have long been used to model acceptability \citep{lau-etal-2017-grammaticality}. For generics, \citet{cilleruelo-etal-2025-generics} introduced ConGen, using surprisal to estimate their implicit quantifiers; \citet{cilleruelo-calderon-etal-2026-generics} later argued that no single quantifier preserves a generic's meaning. However, neither study examines the factors shaping human acceptance. We use surprisal to test whether models reflect these factors and compare their written answers with their expected ratings.

Even when models agree with humans, how they reach their judgments remains unclear. Interpretability research shows that numeric attributes are encoded along linear directions that can shift model outputs when manipulated \citep{heinzerling-inui-2024-monotonic}, and that similar interventions can flip truth judgments across datasets \citep{marks-tegmark-2024-geometry}. Building on \citet{el-shangiti-etal-2025-geometry}, we test whether intervening along directions associated with individual factors shifts model ratings.

\section{Statistical accounts of acceptability}
\label{sec:replication}

\subsection{Experimental design}
We first investigate whether LLMs rely on the same information as people when judging generics, and whether they combine it in the same way. Building on statistical accounts of generic acceptability \citep{ryu-etal-2022-flexible,hoorens-etal-2026-constat}, we base our experimental design on \citet{ryu-etal-2022-flexible}, in which participants rated each generic on two factors. \textbf{Prevalence} is how common people expect a property to be within a kind. For ``Ducks lay eggs,''
participants were asked ``Suppose you encounter a duck X. How likely is it that X lays eggs?''. Ratings range from 0 to 100; where they label the ends 0 (\textit{definitely not}) and 100 (\textit{definitely yes}). \textbf{Cue validity} is how likely something with the property is to belong to the kind, among members of a broader comparison (superordinate) category. Using birds as the
comparison category, participants were asked, ``Suppose a bird X lays eggs. How likely is it that X is a duck?'' Ratings range from $-3$ (\textit{very unlikely}) to $+3$ (\textit{very likely}).
A separate group of participants rated acceptability, from $-3$ (\textit{definitely reject}) to $+3$ (\textit{definitely accept}).

When analyzing their results on human participants, \citet{ryu-etal-2022-flexible} distinguish two types of generics. Some state that one category belongs to another, e.g. ``Dogs are mammals'' (\textbf{IS\_A generics}); others attribute a property to a category, e.g. ``Ducks lay eggs'' (\textbf{feature-describing generics}). For the first type, they find that acceptability depended mainly on prevalence. For the second, both factors mattered and interacted, as in Equation 1:
\begin{equation}
A = -1.81 + 0.04P + 1.16C - 0.01(P \times C) \;\;\mathrm{,}
\end{equation}

where $A$ is acceptability, $P$ prevalence and $C$ cue validity. Properties with high cue validity were accepted even when they were rare (``Mosquitoes carry malaria''), whereas for properties with low cue validity, acceptability depended mainly on prevalence. Refitting this regression on their published item ratings reproduced the coefficients, so we use it as the reference for human judgments.

\label{sec:judgments}
For our experiment, we prompted each model to rate every item for acceptability, prevalence and cue validity with the same question template and scales they used (full prompts in Appendix~\ref{app:prompts}). We then applied the same analyses to each model's ratings: the correlation between prevalence and cue validity; for IS\_A generics, the correlations of acceptability with each factor; for feature-describing generics, the regression of acceptability on prevalence, cue validity and their interaction; and the same correlations within the high- and low-cue-validity groups.

\subsection{Data}

\paragraph{Ryu et al.'s items.}
Our first dataset contains the 84 generics used by \citet{ryu-etal-2022-flexible}, drawn from \citet{prasada-etal-2013-conceptual}: 20 IS\_A and 64 feature-describing generics. Of the latter, 30 have high cue validity (above 1.3) and 34 have lower cue validity. The published mean ratings of acceptability, prevalence and cue validity serve as our human reference, which makes this the dataset on which model and human judgments can be compared directly. They obtained more general comparison categories for cue validity from human participants but did not publish a complete list, so we supplied one for each item using the procedure described for ConGen in Appendix~\ref{app:conversion}. Ours match the examples in their paper, such as ``a bird'' for ``Ducks lay eggs''and ``a toy'' for ``Dolls wear dresses.''

\paragraph{ConGen.}
To test our findings on naturally occurring generics, we constructed a second dataset from the 970 bare generics in ConGen \citep{cilleruelo-etal-2025-generics}. We automatically extracted category and property phrases and assigned comparison categories for the cue-validity questions. An LLM judge checked the conversions, with flagged items reviewed manually. Two annotators reviewed the comparison categories, and a third resolved disagreements (see Appendix~\ref{app:conversion} for details). The final dataset contains 732 items: 582 feature-describing and 150 IS\_A generics. Since human acceptability ratings are unavailable for ConGen, we test whether each model's ratings follow the pattern predicted by \citet{ryu-etal-2022-flexible}.

\subsection{Models}
\label{sec:models}

Our main analyses use six open-weight instruction-tuned language models with 7--9B parameters: Qwen2.5-7B-Instruct~\citep{qwen-2024-qwen25}, Llama-3.1-8B-Instruct~\citep{grattafiori-etal-2024-llama3}, Mistral-7B-Instruct-v0.3~\citep{jiang-etal-2023-mistral}, OLMo-3-7B-Instruct~\citep{olmo-2025-olmo3}, Gemma-2-9B-it~\citep{gemma-2024-gemma2} and Qwen3-8B~\citep{qwen-2025-qwen3}, the latter with disabled reasoning mode, to keep the task closer to the original one for humans. 
To test whether the results generalize, we study two further groups of models. The first contains four larger models that, like the open models above, answer directly: Llama-3.1-70B-Instruct \citep{grattafiori-etal-2024-llama3} and Qwen2.5-72B-Instruct~\citep{qwen-2024-qwen25}, larger versions of two of those models, so that size is the main difference; and two proprietary models of undisclosed size, GPT-4.1~\citep{openai-2025-gpt41} and Gemini-2.5-Flash \citep{google-2026-gemini25}, the latter with reasoning switched off. The second contains two reasoning models, Gemini-3.8-Flash~\citep{google-2026-gemini38flash} and Qwen3.8-Flash~\citep{alibaba-2026-qwen38flash}, a mixture-of-experts model with 125B parameters (6B active per token).

\subsection{Do models follow the human rule?}
\label{sec:results-ryu}

\begin{figure}[t]
\centering
\includegraphics[width=\columnwidth]{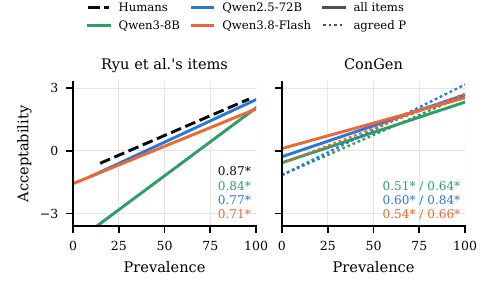}
\caption{\label{fig:isa-qwen}
  IS\_A generics: acceptability against prevalence for humans and one Qwen model per group (fitted
  lines; Pearson $r$, $^{*}$\,$p < .05$). ConGen: all items (solid) and the half on which models agree
  most about prevalence (dotted). All models in Appendix~\ref{app:isa}.}
\end{figure}

We ran each analysis with two sets of predictors. First, we used each model's own prevalence and cue-validity ratings to test whether its acceptability judgments follow the human pattern given its own estimates of these factors. Each model, however, provides a single integer rating per item, whereas human ratings are averaged across participants. Noise in individual ratings may therefore obscure the interaction even when a model combines the factors as people do. To reduce this source of noise, we repeated the analyses using human ratings for Ryu et al.'s items and, for ConGen, ratings averaged within each model's group (7--9B, larger direct-answer or reasoning models). On Ryu et al.'s items, these group averages correlate closely with human ratings ($r = .79$--$.87$ for prevalence and $r = .73$--$.75$ for cue validity; cr. Table~\ref{tab:human-model-corr} in the Appendix for details). We assess whether model judgments follow the human pattern using either the model's own estimates or estimates averaged across participants or models.

\paragraph{IS\_A generics.}
For IS\_A generics, \citet{ryu-etal-2022-flexible} found that acceptability increases with prevalence and decreases with cue validity, while the two factors are negatively correlated. All twelve models reproduce the positive relationship between acceptability and their own prevalence ratings on Ryu et al.'s items, while eleven show a negative correlation between prevalence and cue validity. On ConGen, however, prevalence and cue validity are positively correlated for all but the reasoning models. The relationship between prevalence and acceptability is also weaker for all twelve models. We analyze the reason for the positive correlation between the two factors for ConGen and find that it is mainly a reflection of the difference in dataset construction. Ryu et al.'s items pair a kind with a broader category (\textit{Dogs are mammals}: high prevalence, low cue validity) or reverse that relation (\textit{Mammals are dogs}: low prevalence, high cue validity), placing the factors in opposition by design. ConGen contains few reversed statements and many definitional ones, whose properties are both common within the kind and diagnostic of it.

The weaker relationship between prevalence and acceptability appears to reflect uncertainty about prevalence. Models agree closely on prevalence for Ryu et al.'s items (mean inter-model $r = .92$; correlation between their average and human ratings: $r = .98$), but less so for ConGen ($r = .57$). ConGen includes negations (\textit{Raspberries are not bushes}) and technical descriptions (\textit{Rabbits are hind gut fermenters}) that may make prevalence harder to assess. We take half of ConGen's items with greater agreement among models and find that the correlations between prevalence and acceptability exceed those on the full set for all twelve models, hence suggesting that models rely on prevalence more strongly whenever the prevalence of a property can be clearly assessed. Figure~\ref{fig:isa-qwen} shows the relationship between prevalence and acceptability for one Qwen
model from each group, while Appendix~\ref{app:isa} reports both correlations for all models. We return to this pattern in \S\ref{sec:factors}, where we examine whether prevalence predicts acceptability because it tracks factuality.

\begin{figure*}[t]
\centering
\includegraphics[width=\textwidth]{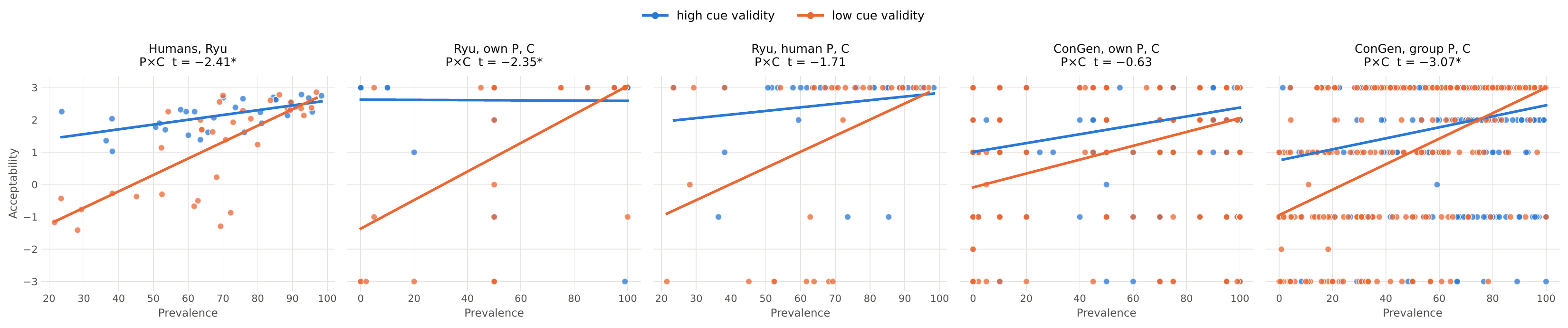}
\caption{\label{fig:feat-interaction}
  Feature-describing generics: acceptability against prevalence (P) for high and low cue validity (C),
  for humans and Qwen3-8B. Converging lines indicate an interaction ($^{*}$\,$p < .05$). Human
  ratings are means over participants; the model's are single integer ratings, hence the discrete
  values. All models in Table~\ref{tab:feat-interaction}.}
\end{figure*}

\paragraph{Feature-describing generics.}
For feature-describing generics, \citet{ryu-etal-2022-flexible} found that people weigh prevalence and cue validity against each other. Both make a generic more acceptable, but the higher the cue validity, the less the prevalence matters. Models are sensitive to both factors and follow the human pattern in both datasets. On ConGen, prevalence and cue validity both predict acceptability in nearly all models. On Ryu et al.'s items, both factors again predict acceptability, more strongly with the human ratings of the two factors than with the models' own, but the interaction is weak: it usually points in the human direction but is rarely significant for individual models. Averaged ratings, whether from people or models show the human pattern clearly.

Figure~\ref{fig:feat-interaction} illustrates this for Qwen3-8B, and the same pattern recurs across models (Appendix~\ref{app:feat-interaction}). As for humans, cue validity matters most when prevalence is low, and its effect weakens as prevalence increases. The interaction is clearer in the larger models, and averaging ratings across models brings it out in the smaller ones as well; with group-averaged ratings on ConGen, most models combine the two factors significantly in the same way as
humans. Overall, the models apply the human rule in its direction, and the larger ones come close to its strength. The interaction is about as strong on Ryu et al.'s items as on ConGen, but with nine times as many items, it reaches significance for many more models on ConGen, suggesting that its weakness on Ryu et
al.'s items is due to the small sample size.

\section{Beyond prevalence and cue validity}
\label{sec:factors}

Prevalence and cue validity explain only part of the variance in acceptability ratings (55\% of the variance in people's ratings and 3--42\% in each model's own ratings on Ryu et al.'s feature-describing items). We therefore examined four further factors that might account for how models judge generics.


In the psychology of generics, two further factors explain judgments that prevalence cannot. Some generics are rejected although almost every member has the property, as in \emph{Humans are over three years old}. \citet{prasada-etal-2013-conceptual} account for this with \emph{characteristicness}: a generic is accepted when the property is characteristic of the kind itself, not merely common among its members. Other generics are accepted although only a few members have the property, as in \emph{Mosquitoes carry malaria}. \citet{leslie-2008-generics} and \citet{cimpian-etal-2010-generic} attribute this to \emph{strikingness}: a dangerous or otherwise remarkable property licenses a generic on little evidence. Thus, both are candidates to be additional factors shaping the ratings of the models.

The second source is the models themselves. To see whether they attend to considerations that the accounts of human judgment leave out, we asked each 7--9B model to explain its reasoning before rating Ryu et al.'s items, and compared the words in explanations that ended in rejection with those that ended in acceptance (Appendix~\ref{app:cot}). Two kinds of consideration stood out. Explanations ending in acceptance appealed to whether the statement is factually correct, and those ending in rejection objected that it is inaccurate or overgeneralizes, applying to more members than it should. We therefore added two new factors: \emph{factuality}, the model’s judgment of whether the statement is accurate; and \emph{overreach}, its judgment of whether the statement claims more than is warranted. Overreach connects to two earlier findings: people tend to accept universal claims such as \emph{All ducks lay eggs} when the corresponding generic is true \citep{leslie-etal-2011-ducks}, and language models overgeneralize from generics in a similar way \citep{allaway-etal-2024-exceptions}. 

Each model rated every ConGen and Ryu et al.'s item on the four factors, with questions built like those for prevalence and cue validity (Appendix ~\ref{app:prompts}). We test the factors separately for IS\_A and feature-describing generics, since \citet{ryu-etal-2022-flexible} found different rules for the two on humans. To test what the new factors contribute, we needed a baseline built from the two established ones. We use  $A \sim P + C$  as the baseline for both types, since adding their interaction rarely improves held-out predictions (Appendix~\ref{app:baselines}).

\subsection{Factors from the psychology of generics}
\label{sec:psych-factors}

We first test whether either factor restates prevalence or cue validity. Across the twelve models and both datasets, no correlation of characteristicness or strikingness with a model's prevalence or cue validity exceeds .63 in absolute value. Both relate to prevalence in the direction the psychological accounts predict. Characteristicness rises with prevalence in every model and dataset and strikingness falls with it in all but one case (Appendix \ref{app:psych-overlap}; Figure~\ref{fig:psych-overlap}). The characteristicness ratings also capture what this factor describes. Ryu et al.'s items come from \citet{prasada-etal-2013-conceptual}, who sorted them into seven types, separating properties that are part of what the kind is (\emph{Ducks lay eggs}) from ones that are merely common (\emph{Cars have radios}) or striking (\emph{Mosquitoes carry malaria}). Every model rates the first group as more characteristic, and the models' ranking of the seven types follows people's judgments in \citeauthor{prasada-etal-2013-conceptual}'s study (Appendix~\ref{app:validate}).


Within each generic type, we add both factors to the baseline and standardise all ratings, so that each factor is credited only with what it explains beyond the others, then compare how much of the variance the model explains with and without them.

\begin{figure}[t]
\centering
\includegraphics[width=\columnwidth]{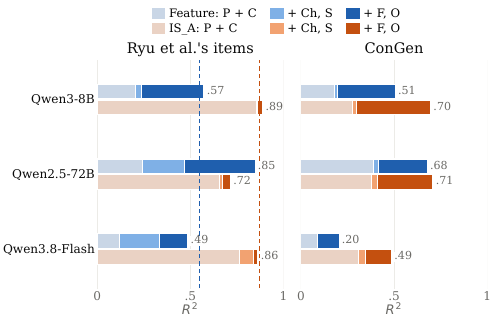}
\caption{\label{fig:factor-summary}
  $R^2$ of acceptability as factors are added (one Qwen per group); dashed: people's P + C.
  All models in Appendix~\ref{app:factor-tables}.}
\end{figure}

\paragraph{Characteristicness.}
Characteristicness raises acceptability in almost every model, for both types of generic (Figure~\ref{fig:factor-summary}). Its effect is largest where the baseline does worst. On Ryu et al.’s feature-describing items, prevalence and cue validity explain little of the ratings of most larger and reasoning models (3--24\% of the variance, against 55\% of people’s), but with characteristicness and strikingness added they explain 33--50\%. On ConGen the effect is consistent but small. Characteristicness also accounts for items the statistical factors get wrong: \emph{Canadians are right-handed} has high prevalence, yet most larger models reject it, as people do, and nearly all models rate the property as not characteristic of Canadians. More generally, the baseline’s errors track characteristicness in most models for both types. The exception is Ryu et al.’s 20 IS\_A items, where prevalence and cue validity already explain most of the variance and the factors are too closely related to separate (Appendix~\ref{app:psych-isa}).

\paragraph{Strikingness.}
Strikingness, in contrast, adds little for either type of generic. Its weights are close to zero in most models, and where they are not, they mostly point the wrong way. This is not a failure of measurement, since every model recognises the striking generics as the most striking (Appendix~\ref{app:validate}). The models know which properties are striking but do not treat strikingness as a reason to accept a generic, nor does it matter more for rare properties, as \citet{leslie-2008-generics} predicts (Appendix~\ref{app:psych-interactions}). People may not either: \citet{hermans-etal-2026-justifiable} found that dangerousness affected how justifiable people found generics only at intermediate prevalence. On ConGen, most models find generics about striking properties less acceptable, but only because such properties are rare. They treat them like any other generic about a rare property.

Taken together, the two factors behave very differently. The models attend to characteristicness beyond prevalence and cue validity, most clearly in the larger and reasoning models, but they do not use strikingness. Yet even with both factors, much of the variance in the ratings remains unexplained, especially on ConGen’s natural generics. 
Next, we analyze the factors coming from the self-explanations of the models.

\subsection{Factors from the models' justification}
\label{sec:model-factors}

Adding factuality to the four psychological factors has the largest weight in almost every model, for both types of generic and on both datasets, and together with overreach it raises the variance explained significantly in every model (Figure \ref{fig:factor-summary}; all models in Appendix~\ref{app:factor-tables}). The gain is largest where the psychological factors explain least: on ConGen, factuality often doubles the variance explained. Factuality is not the same as acceptability, though. A statement can be true of most members of a kind and still make a poor generic, as with \emph{Humans are over three years old} \citep{leslie-etal-2011-ducks}, and the models make this distinction too: they reject up to a quarter of the statements they judge accurate, and on ConGen factuality alone often accounts for less than half of their acceptability ratings. Overreach plays a smaller role. Where it matters, it lowers acceptability, mainly on ConGen and in the larger models, and in several models it adds nothing once factuality is known.

However, factuality and overreach are not independent of the psychological factors. Prevalence, cue validity, characteristicness and strikingness explain a substantial part of the models’ factuality ratings: statements about properties that most members have, or that are characteristic of the kind, are judged more accurate and less overreaching in almost every model. As a result, almost everything the psychological factors explain about acceptability is also explained by factuality and overreach (Appendix~\ref{app:variance-partition}). The human factors thus still matter for the models, but they reach acceptability largely through judgments of whether a statement is true and whether it claims too much. On top of this shared part, factuality and overreach explain about as much again of their own. Appendix~\ref{app:divergence} discusses individual generics on which people and models depart from the statistical rule, and what the models’ factor ratings suggest about why.

\subsection{Signal from probabilistic measures}
Everything so far rests on what the models answer when asked. Previous NLP research showed that responses to prompts can understate what a model knows, and measures read directly from its probabilities are often more reliable \citep{hu-levy-2023-prompting, kauf-etal-2024-log}. We therefore also measured acceptability without asking a question: we gave each 7--9B model only the bare generic and read off how probable it finds the property given the kind, as the average log-probability of \emph{carry malaria} after \emph{Mosquitoes} \citep[P-Acceptability;][]{cilleruelo-etal-2025-generics}. For generics, this signal agrees worse with the ratings of people than the models' answers do, probably because the generics differ in every word, so their probabilities also reflect word frequency, word length, and how often such statements are written \citep{kauf-etal-2024-log}. Still, with people's prevalence and cue validity as predictors, it shows the same pattern as in Section~\ref{sec:replication}: IS\_A generics become more probable with prevalence, feature-describing ones with cue validity, and the interaction points in the human direction in every model, though too weakly to be significant (Appendix~\ref{app:surprisal}). Following \citet{cilleruelo-etal-2025-generics}, we also placed \emph{All}, \emph{Most} or \emph{Some} before each generic and recorded which version makes the property most probable. This best-fitting quantifier tracks prevalence, with \emph{All} for the most prevalent properties and \emph{Some} for the least. Whether we ask the models or read their probabilities, both show which factors go together with acceptability, not which ones the model uses to reach it. We therefore look inside the models, testing whether moving their hidden states along each factor’s direction changes their judgment.

\section{Do factor directions change the models' judgments?}
\label{sec:exp2}
\begin{table*}
\centering
\small
\begin{tabular*}{\textwidth}{@{\extracolsep{\fill}}llrrrrrrr@{}}
\toprule
\textbf{Model} & \textbf{Layer} & F & P & C & Ch & O & S & Random \\
\midrule
Qwen2.5-7B & 15 / 28 & \textbf{.72}$^{*}$ & .51$^{*}$ & .30$^{*}$ & .18$^{*}$ & \textbf{\boldmath$-$.11}$^{*}$ & .22$^{*}$ & $-$.06$^{*}$ \\
Llama-3.1-8B & 13 / 32 & \textbf{1.88}$^{*}$ & 1.23$^{*}$ & 1.65$^{*}$ & 1.52$^{*}$ & \textbf{\boldmath$-$.16}$^{*}$ & .48$^{*}$ & $-$.09$^{*}$ \\
Mistral-7B & 13 / 32 & \textbf{2.07}$^{*}$ & 1.82$^{*}$ & 1.81$^{*}$ & 1.45$^{*}$ & \textbf{\boldmath$-$.88}$^{*}$ & .51$^{*}$ & .05$^{*}$ \\
OLMo-3-7B & 13 / 32 & \textbf{2.50}$^{*}$ & 1.42$^{*}$ & 1.44$^{*}$ & .89$^{*}$ & \textbf{\boldmath$-$.57}$^{*}$ & .39$^{*}$ & .11$^{*}$ \\
Gemma-2-9B & 17 / 42 & \textbf{.59}$^{*}$ & .47$^{*}$ & \textbf{.59}$^{*}$ & .35$^{*}$ & \textbf{\boldmath$-$.64}$^{*}$ & .36$^{*}$ & .01 \\
Qwen3-8B & 20 / 36 & \textbf{1.22}$^{*}$ & 1.10$^{*}$ & 1.00$^{*}$ & 1.11$^{*}$ & \textbf{\boldmath$-$.57}$^{*}$ & .43$^{*}$ & .07$^{*}$ \\
\bottomrule
\end{tabular*}
\caption{\label{tab:steering}
  Change in expected acceptability per unit of push along each factor's direction, at factuality's
  peak layer. Bold: largest effect and overreach; $^{*}$\,beyond two standard errors.}
\end{table*}

We examined the internal states of the six 7--9B models, adapting the method of \citet{el-shangiti-etal-2025-geometry}. For each factor, we find a direction in the model's hidden states and ask whether moving the hidden state along that direction changes the model's acceptability judgment.

\subsection{Method}
For each factor, we recorded the model's hidden states while it read the question about that factor, for the feature-describing items in ConGen. At each layer, we identified the factor's direction by regressing the hidden states on the model's ratings of all six factors, so that each direction captures one factor with the others held constant. As ratings, we used the model's expected rating, the average of the possible answers weighted by their probabilities, rather than the number it wrote, because expected ratings are continuous and read from the same input as the hidden states. Fitting to the written ratings gives essentially the same result (Appendix~\ref{app:written-labels}), and the two kinds of rating agree closely, though not perfectly (Appendix~\ref{app:faithfulness}).

While the model reads the acceptability prompt, we added a factor's direction to its hidden state at one layer, in doses of $\pm 0.5$ and $\pm 1$ times the typical distance of the states from their mean, and measured the change in its expected acceptability rating. A random direction pushed by the same amount served as a control. Interventions were run on 200 items per model, at layers between 40\% and 85\% of the model's depth. We read and changed the hidden state at the last token of the question, before the instruction that specifies the answer format. Five of the factor questions share the acceptability scale, so where the model is about to answer, a factor's direction partly encodes the number it will produce, and pushing along any such direction raises the rating regardless of what the factor means. Because models process text from left to right, the hidden state at the end of the question cannot contain information about the answer format.

Steering moves every item by the same, fixed amount. To test whether the differences that actually exist between items matter, we also swapped positions along each direction. Items in the bottom quarter of a factor's ratings were given the position along its direction of a randomly chosen item from the top quarter, leaving the rest of the hidden state unchanged, and vice versa. We report the change in expected acceptability for the items moved up minus that for the items moved down, at the same layer and token as the steering.

\subsection{Results}
Pushing along each factor's direction changes the acceptability rating, in the direction the factor's meaning predicts, and more than pushing along a random direction of the same size, in almost all cases by far (Table~\ref{tab:steering}). Factuality has the largest positive effect in every model, tied with cue validity in Gemma. Prevalence, cue validity and characteristicness follow, in varying order, and strikingness, which plays almost no role in behavior, has a smaller but clear effect. Notably, cue validity, and in some models characteristicness, both of which carry little unique weight once factuality is in the behavioral model, change the judgment about as much as prevalence, so their small behavioral weights may reflect information shared with factuality rather than a lack of use. Overreach shows that the directions carry what a factor means rather than a tendency to give a higher number. Its question uses the same scale as acceptability, but a high value means that a statement claims too much, so a direction encoding only a higher answer would raise acceptability, whereas one encoding overreach should lower it. Pushing along it lowers acceptability in every model (Table~\ref{tab:steering}), and giving items the position of overreaching ones lowers their rating in five of the six models (Appendix~\ref{app:swaps}).

Swapping the positions of items along a factor's direction shows which of these differences matter at the size they actually take (Appendix~\ref{app:swaps}). Giving items the position of items rated more factual raises their rating by up to 1.8 points in five of the six models, and the corresponding swaps raise it for prevalence and cue validity in every model and lower it for overreach in five of the six. Swaps along strikingness shift it only slightly, by at most .07 points, and swaps along the random direction leave it unchanged. At realistic sizes, factuality thus matters most and strikingness barely at all, in line with their roles in behavior, while strikingness has a clear effect only when pushed far beyond the differences between items. 
\section{Conclusion and future work}

Psycholinguistic factors explain part of how LLMs judge generics. The models combine prevalence and cue validity as the human rule predicts, though less clearly, and favor characteristic properties but, not striking ones. Above all, they judge whether a generic is true, and the human factors shape their judgments largely through that. Inside the models, each factor has a direction that moves the judgment as its meaning predicts. Although the factors explain much of how LLMs judge generics, a large share of the variance still remains unexplained.

Future work should identify the factors behind the unexplained variance, especially for naturally occurring generics, and collect human acceptability ratings for them so that models and people can be compared on more items. It should also trace how models combine  factors beyond single directions and test generics in context, in other languages, and about social groups, where overgeneralizing may cause the most harm \citep{peters2024generics}.

\section*{Limitations}

Human ratings exist only for \citet{ryu-etal-2022-flexible}'s 84 items, so the comparison with people on ConGen is indirect as we test whether models follow \citet{ryu-etal-2022-flexible}'s human pattern, not whether they agree with people on natural generics, and the small item set limits power for individual models. 

All factors are elicited from the models with one prompt each and, in the behavioural analyses, a single integer rating; apart from prevalence and cue validity, the questions are our own operationalizations, and, to the best of our knowledge, no human ratings with them exist, only \citet{prasada-etal-2013-conceptual}'s related ratings of
principled connections. 

The internal analysis covers only the six 7--9B open-weight models, assumes a single linear direction per factor, and intervenes at one layer and token position on feature-describing ConGen items. 

All items are in English, presented out of context, so the results may not carry over to other languages, to generics in discourse, or to generics about social groups.

\section*{Ethics and broader impact}

We collected no new human-subject data. We use only published resources: the ratings of \citet{ryu-etal-2022-flexible} and the ConGen dataset of \citet{cilleruelo-etal-2025-generics}, under their respective licences, together with openly
released and commercially available language models used within their terms of use. The comparison categories were reviewed by the authors, so no external annotators were involved. 

Generics can convey and reinforce stereotypes, and naturally occurring generics may
include statements about social groups; we analyze such statements only as linguistic data and do not
endorse their content.  Our findings caution against treating LLMs' agreement with people as evidence of
human-like reasoning, and against using LLMs as substitutes for human participants without checking their judgments. The experiments were run on NVIDIA A100 GPUs (80 GB) for the open models and through paid APIs for the proprietary ones. AI assistants helped with coding and editing; the authors designed the study and checked all code, results and wording.

\bibliography{custom}
\clearpage
\appendix


\section*{Appendix}
\section{Prompts}
\label{app:prompts}

Each question was sent as a single user message in the model's chat format, with no system message. For Qwen3, \texttt{/no\_think} was added on a new line after the question and thinking was switched off in the chat template. Table~\ref{tab:prompts} gives all seven prompts: Ryu et al.'s three questions and the four additional factors. The placeholders in angle brackets were filled from each item's materials: the generic sentence, the category (e.g.\ \textit{a lion}), the feature (\textit{has a mane}) and the comparison category (\textit{an animal}).

\begin{table*}
\centering
\small
\begin{tabular}{@{}>{\raggedright\arraybackslash}p{0.15\textwidth}
                >{\raggedright\arraybackslash}p{0.81\textwidth}@{}}
\toprule
\textbf{Factor} & \textbf{Prompt} \\
\midrule
Acceptability &
\texttt{Please rate how acceptable the following sentence is: "<generic>"}\newline\newline
\texttt{Answer with a single number from -3 to +3, where -3 = "Definitely reject", 0 = "Not sure", and +3 = "Definitely accept." Respond with only the number, nothing else.} \\
\addlinespace[6pt]
Prevalence &
\texttt{Suppose you encounter <category> X. How likely is it that X <feature>?}\newline\newline
\texttt{Answer with a single number from 0 to 100, where 0 = "definitely not" and 100 = "definitely yes." Respond with only the number, nothing else.} \\
\addlinespace[6pt]
Cue validity &
\texttt{Suppose <comparison category> X <feature>. How likely is it that X is <category>?}\newline\newline
\texttt{Answer with a single number from -3 to +3, where -3 = "Very unlikely", 0 = "Not sure", and +3 = "Very likely." Respond with only the number, nothing else.} \\
\midrule
Factuality &
\texttt{How factually accurate is the following statement?}\newline\newline
\texttt{"<generic>"}\newline\newline
\texttt{Answer with a single number from -3 to +3, where -3 = "factually wrong", 0 = "Not sure", and +3 = "factually correct." Respond with only the number, nothing else.} \\
\addlinespace[6pt]
Overreach &
\texttt{Sentences that generalise about a kind vary in how strong a claim they make. Some say no more than the facts support; others assert more about the kind than is warranted.}\newline\newline
\texttt{Does the following sentence claim more than is warranted?}\newline\newline
\texttt{"<generic>"}\newline\newline
\texttt{Answer with a single number from -3 to +3, where -3 = "claims no more than is warranted", 0 = "Not sure", and +3 = "claims far more than is warranted." Respond with only the number, nothing else.} \\
\addlinespace[6pt]
Characteristicness &
\texttt{Some properties are characteristic of a kind: the kind is disposed to have them, whether or not every member does. Others are merely common among members without being characteristic of the kind itself.}\newline\newline
\texttt{Is "<feature>" a characteristic property of <category>?}\newline\newline
\texttt{Answer with a single number from -3 to +3, where -3 = "not characteristic at all, just something many happen to have", 0 = "Not sure", and +3 = "strongly characteristic of the kind." Respond with only the number, nothing else.} \\
\addlinespace[6pt]
Strikingness &
\texttt{Suppose you encounter <category> X, and you learn that X <feature>.}\newline\newline
\texttt{How alarming, dangerous or noteworthy would that be?}\newline\newline
\texttt{Answer with a single number from -3 to +3, where -3 = "completely unremarkable", 0 = "Not sure", and +3 = "highly alarming or noteworthy." Respond with only the number, nothing else.} \\
\bottomrule
\end{tabular}
\caption{\label{tab:prompts}
  Prompts for Ryu et al.'s three questions (top) and the four additional factors (bottom).
  Placeholders in angle brackets are filled from each item's materials; a line break marks a blank
  line in the prompt.}
\end{table*}
\section{Constructing the ConGen items}
\label{app:conversion}

\paragraph{Parsing.}
We parsed each sentence with spaCy~\citep{honnibal-etal-2020-spacy} and extracted the subtree rooted at the main verb's nominal subject as the category phrase. The head noun was singularised and given an indefinite article (``Millipedes'' $\rightarrow$ ``a millipede''), except for nouns identified as uncountable, such as ``furniture.'' A noun was treated as uncountable if its singular and plural forms were identical or if its plural was much less frequent than its singular in a word-frequency list~\citep{speer-2022-wordfreq}. The remaining clause formed the property phrase, with the verb inflected for a third-person singular subject (``eat'' $\rightarrow$ ``eats''; ``do not grow'' $\rightarrow$ ``does not grow''). Following \citet{ryu-etal-2022-flexible}, we labelled items as IS\_A when the root was a copula with a noun phrase complement (``Joshua trees are desert plants''), and as feature-describing otherwise.

\paragraph{Rejections and repairs.}
A sentence was rejected only if no item could be built from it: its subject was a singular proper noun, which refers to an individual rather than a kind; it had no identifiable subject; its property phrase became empty after repair; its subject was a countable singular noun, which does not express a generic; or its subject was not a noun. Otherwise, sentences were repaired where needed. This produced 868 items, 120 of which had been repaired.

\paragraph{Deriving comparison categories.}
We used WordNet~\citep{miller-1995-wordnet} to identify a comparison category for each item. To select the sense of the category noun, we embedded the generic and its phrases, along with each candidate sense's gloss, using \texttt{all-MiniLM-L6-v2}, a Sentence-Transformers model \citep{reimers-gurevych-2019-sentence} based on MiniLM \citep{wang-etal-2020-minilm}. We selected the sense with the highest cosine similarity only if it exceeded the similarity of the most frequent sense by more than 0.20; otherwise, we retained the most frequent sense. We then searched all hypernym paths from the selected sense for the nearest ancestor meeting two criteria. First, it needed a SemCor \citep{miller-etal-1993-semantic} sense frequency of at least 5, to favour familiar categories over technical terms such as ``anseriform bird.'' Second, it needed at least 100 descendant synsets, to exclude categories that were too narrow, such as ``educator'' for ``teacher.'' Candidates were ranked by distance and then by SemCor frequency. For IS\_A generics, the comparison category also had to be strictly broader than the category named in the property phrase.

\paragraph{Verifying the conversion.}
An LLM judge (Gemini 2.5 Pro \citep{google-2026-gemini25}) assessed whether each generated prevalence question preserved the meaning of the original sentence. It received the original sentence and the generated question, and returned a verdict and an error category, without being told whether the item had been repaired. The authors reviewed the 253 flagged items, correcting category or property phrases, shortening sentences or removing items. This left 766 items.

\paragraph{Reviewing comparison categories.}
The same judge assessed whether each comparison category captured the alternatives people would consider for the target kind, following \citet{ryu-etal-2022-flexible}, for example, ``a toy'' for ``Dolls wear dresses.'' It approved 209 categories. Two annotators independently reviewed the remaining 557, either retaining the proposed category, suggesting an alternative or requesting that the item be removed. The annotators agreed on 394 items. A third annotator resolved disagreements by choosing between options presented anonymously and in random order, without being told that they arose from a disagreement. The final dataset contained 732 items. 

\section{Detailed results: statistical measures}

\subsection{Agreement of model and human ratings}
\label{app:human-model-corr}

Table~\ref{tab:human-model-corr} shows how closely each model's ratings, and its group's averaged ratings, track the human means on Ryu et al.'s feature-describing generics. Individual models vary widely: their prevalence ratings correlate with people's at $r = .38$--$.85$ and their cue-validity ratings at $r = .34$--$.77$, with OLMo-3-7B lowest on prevalence and Llama-3.1-8B on cue validity. Averaging within a group reduces this variation. The group averages correlate with human prevalence at $r = .79$--$.87$ and with human cue validity at $r = .73$--$.75$, matching or exceeding the best single model in each group in all but one case. This is why, on ConGen, where human ratings are unavailable, we also test the interaction with group-averaged prevalence and cue validity (\S\ref{sec:results-ryu}).

\begin{table}
\centering
\small
\setlength{\tabcolsep}{4pt}
\begin{tabular}{llrrr}
\toprule
\textbf{Group} & \textbf{Model} & A & P & C \\
\midrule
\multirow{7}{*}{Open, 7--9B} & Qwen2.5-7B & .74 & .75 & .66 \\
 & Mistral-7B & .62 & .69 & \textbf{.73} \\
 & OLMo-3-7B & .47 & .38 & .41 \\
 & Gemma-2-9B & \textbf{.83} & \textbf{.79} & .69 \\
 & Qwen3-8B & .60 & .61 & .54 \\
 & Llama-3.1-8B & .72 & .65 & .34 \\
 & \textit{Group average} & .77 & \textbf{.79} & \textbf{.73} \\
\midrule
\multirow{5}{*}{\shortstack[l]{Large,\\direct answer}} & Llama-3.1-70B & .77 & .74 & .69 \\
 & Qwen2.5-72B & .82 & .78 & \textbf{.77} \\
 & GPT-4.1 & .79 & .84 & .70 \\
 & Gemini-2.5-Flash & .66 & .81 & .48 \\
 & \textit{Group average} & \textbf{.84} & \textbf{.87} & .75 \\
\midrule
\multirow{3}{*}{Reasoning} & Gemini-3.8-Flash & .68 & .84 & .68 \\
 & Qwen3.8-Flash & .58 & .85 & .71 \\
 & \textit{Group average} & \textbf{.71} & \textbf{.87} & \textbf{.73} \\
\bottomrule
\end{tabular}
\caption{\label{tab:human-model-corr}
  Pearson correlations of each model's ratings of acceptability (A), prevalence (P) and cue validity
  (C) with the human means, over Ryu et al.'s 64 feature-describing generics. Group average: the
  ratings averaged over the models of the group. Bold: highest in each group.}
\end{table}

\subsection{IS\_A generics: all models}
\label{app:isa}

Table~\ref{tab:isa-agree} gives, for every model, the correlations summarised in \S\ref{sec:results-ryu}: prevalence with cue validity, and prevalence with acceptability, on Ryu et al.'s items and on ConGen, with ConGen split by how much the models agree on prevalence. 

\begin{table*}
\centering
\small
\begin{tabular*}{\textwidth}{@{\extracolsep{\fill}}llrr@{\hspace{3em}}rrrr@{}}
\toprule
 & & \multicolumn{2}{c@{\hspace{3em}}}{$r_{P,C}$} & \multicolumn{4}{c@{}}{$r_{A,P}$} \\
\cmidrule(r{3em}){3-4} \cmidrule{5-8}
 & & & & & & \multicolumn{2}{c@{}}{ConGen, prevalence} \\
\cmidrule{7-8}
\textbf{Group} & \textbf{Rater} & Ryu & ConGen & Ryu & ConGen & agreed & disputed \\
\midrule
Humans & \citet{ryu-etal-2022-flexible} & $-$.64$^{*}$ & -- & .87$^{*}$ & -- & -- & -- \\
\midrule
\multirow{6}{*}{\begin{tabular}[c]{@{}l@{}}Open,\\7--9B\end{tabular}} & Qwen3-8B & $-$.68$^{*}$ & \textbf{.24}$^{*}$ & \textbf{.84}$^{*}$ & .51$^{*}$ & .64$^{*}$ & .37$^{*}$ \\
 & Gemma-2-9B & $-$.72$^{*}$ & .29$^{*}$ & \textbf{.84}$^{*}$ & \textbf{.59}$^{*}$ & \textbf{.75}$^{*}$ & \textbf{.42}$^{*}$ \\
 & Mistral-7B & \textbf{\boldmath$-$.63}$^{*}$ & .35$^{*}$ & .82$^{*}$ & .37$^{*}$ & .62$^{*}$ & .08 \\
 & Llama-3.1-8B & .10 & .35$^{*}$ & .81$^{*}$ & .44$^{*}$ & .47$^{*}$ & .40$^{*}$ \\
 & Qwen2.5-7B & $-$.68$^{*}$ & .32$^{*}$ & .62$^{*}$ & .50$^{*}$ & .64$^{*}$ & .35$^{*}$ \\
 & OLMo-3-7B & $-$.47$^{*}$ & .31$^{*}$ & .61$^{*}$ & .37$^{*}$ & .66$^{*}$ & .00 \\
\midrule
\multirow{4}{*}{\begin{tabular}[c]{@{}l@{}}Large,\\direct\\answer\end{tabular}} & Llama-3.1-70B & $-$.70$^{*}$ & .35$^{*}$ & \textbf{.80}$^{*}$ & .49$^{*}$ & .72$^{*}$ & .29$^{*}$ \\
 & Qwen2.5-72B & \textbf{\boldmath$-$.59}$^{*}$ & .39$^{*}$ & .77$^{*}$ & \textbf{.60}$^{*}$ & \textbf{.84}$^{*}$ & .38$^{*}$ \\
 & Gemini-2.5-Flash & $-$.70$^{*}$ & .27$^{*}$ & .74$^{*}$ & .59$^{*}$ & .72$^{*}$ & \textbf{.47}$^{*}$ \\
 & GPT-4.1 & $-$.73$^{*}$ & \textbf{.25}$^{*}$ & .74$^{*}$ & .56$^{*}$ & .72$^{*}$ & .38$^{*}$ \\
\midrule
\multirow{2}{*}{Reasoning} & Gemini-3.8-Flash & \textbf{\boldmath$-$.71}$^{*}$ & \textbf{\boldmath$-$.17}$^{*}$ & \textbf{.78}$^{*}$ & .38$^{*}$ & .49$^{*}$ & .28$^{*}$ \\
 & Qwen3.8-Flash & $-$.72$^{*}$ & .01 & .71$^{*}$ & \textbf{.54}$^{*}$ & \textbf{.66}$^{*}$ & \textbf{.43}$^{*}$ \\
\bottomrule
\end{tabular*}
\caption{\label{tab:isa-agree}
  IS\_A generics: correlation of prevalence (P) with cue validity (C) and with acceptability (A), each
  rater's own ratings. Agreed / disputed: the 75 ConGen items on which the twelve models' prevalence
  ratings vary least / most (split at the median standard deviation, 22.8 points). The models'
  prevalence ratings agree closely on Ryu et al.'s items (mean inter-model $r = .92$; model average
  vs.\ human ratings, $r = .98$) but less on ConGen's ($r = .57$). Bold: within each group, the model
  closest to the human value reported by \citet{ryu-etal-2022-flexible} ($r_{P,C} = -.64$, $r_{A,P} = .87$).
  Ryu et al.: 20 items; ConGen: 150. $^{*}$\,$p < .05$.
}
\end{table*}

\subsection{The prevalence $\times$ cue validity interaction}
\label{app:feat-interaction}

Table~\ref{tab:feat-interaction} reports the interaction for every model under each analysis described in \S\ref{sec:results-ryu}. Taken together, the estimates show a consistent direction. Of the 48 estimates obtained with linear models (twelve models under four analyses), 43 are negative, in the direction of the human interaction, and 14 are significantly so; none is significantly positive. The models therefore do not combine prevalence and cue validity in the opposite way to humans; they differ in how clearly they show the human pattern.

\begin{figure*}[p]
\centering
\includegraphics[width=\textwidth]{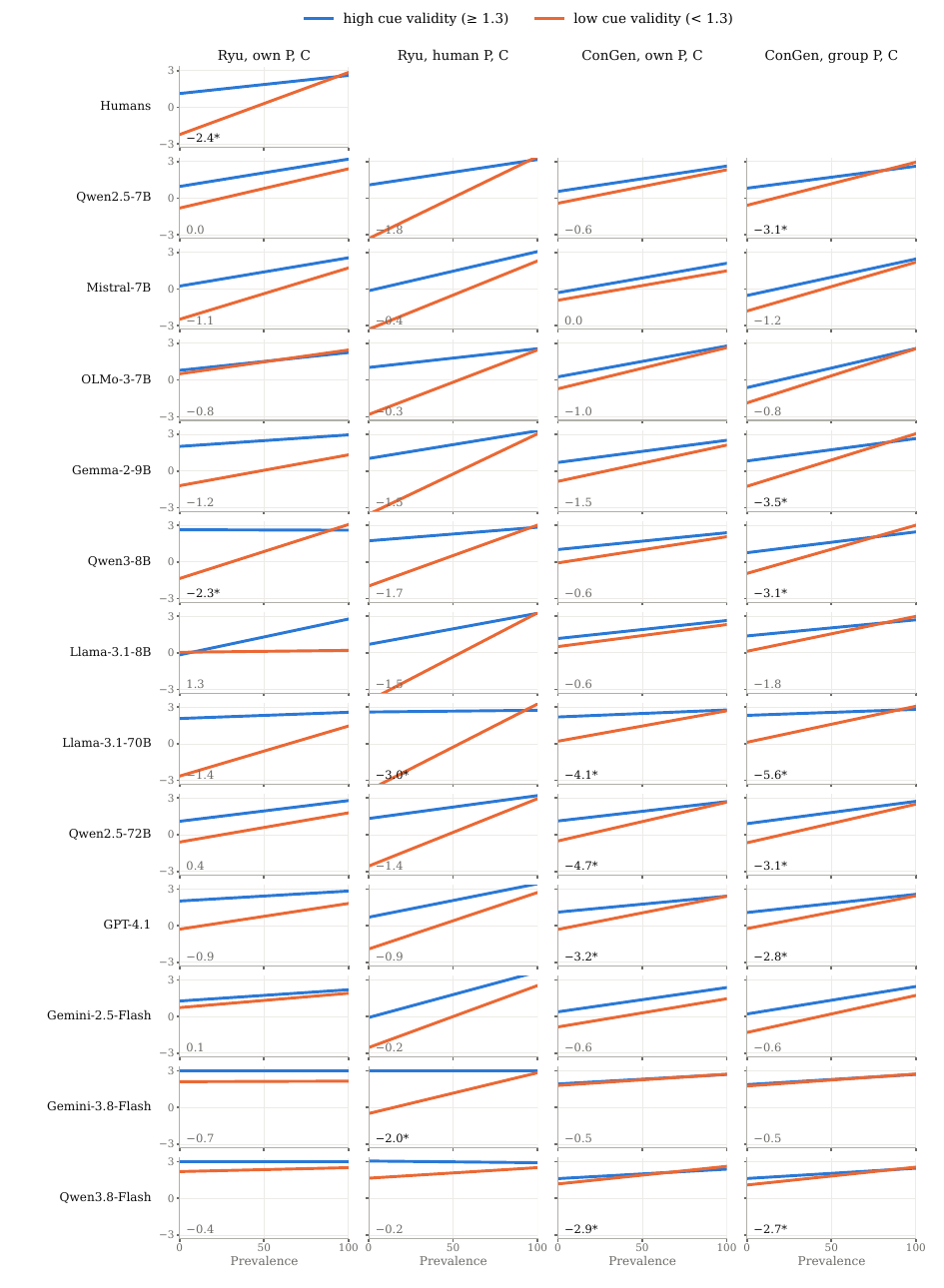}
\caption{\label{fig:feat-interaction-all}
  Feature-describing generics: acceptability against prevalence for high and low cue validity, for
  every model and analysis (fitted lines; $t$ of the P $\times$ C interaction, $^{*}$\,$p < .05$).
  Converging lines: the human interaction.}
\end{figure*}

The same consistency appears in the shape of the relationship (Figure~\ref{fig:feat-interaction-all}). For each model and analysis, we fitted acceptability against prevalence separately for properties with high and low cue validity. With the human ratings of prevalence and cue validity, and with the ratings averaged over each group of models, the advantage of a high cue validity is largest at low prevalence and smallest at high prevalence (negative interaction) for all twelve models, though not always significant. Against the human ratings, the interaction is negative for every model, as for humans, although it is weaker than in humans for most models, and significant only for the two models where it is stronger (Llama-3.1-70B and Gemini-3.8-Flash). With the models’ own ratings the same shape appears in a flatter form; it is nearly flat for Mistral-7B on ConGen and Gemini-3.8-Flash on Ryu et al.’s items, and reversed only for Llama-3.1-8B on Ryu et al.’s items.


Whether an individual model shows a significant interaction depends on the data and on the noise in its ratings. On Ryu et al.'s 64 feature-describing items, few models reach significance, and which ones do is unrelated to which do so on ConGen: the models' estimates on the two datasets are uncorrelated. The small item set offers too little power to rank models, which is why ConGen provides the more informative test. Noise in the models' own ratings matters mainly for the smaller models. Averaging prevalence and cue validity within each group strengthens the interaction for five of the six 7--9B models, and makes it significant for three, but leaves the larger models largely unchanged. The judgments of the smaller models may thus already reflect the interaction, but their own answers about the two factors are too noisy to reveal it, whereas the larger models' answers are consistent enough with their judgments for the interaction to appear without averaging.

\begin{table*}
\centering
\small
\begin{tabular*}{\textwidth}{@{\extracolsep{\fill}}llrr@{\hspace{3em}}rrr@{}}
\toprule
 & & \multicolumn{2}{c@{\hspace{3em}}}{Ryu et al.'s items} & \multicolumn{3}{c@{}}{ConGen} \\
\cmidrule(r{3em}){3-4} \cmidrule{5-7}
\textbf{Group} & \textbf{Rater} & own & human P, C & own & group avg.\ P, C & own, ordinal \\
\midrule
Humans & \citet{ryu-etal-2022-flexible} & \textbf{\boldmath$-$.166}$^{*}$ & -- & -- & -- & -- \\
\midrule
\multirow{6}{*}{\begin{tabular}[c]{@{}l@{}}Open,\\7--9B\end{tabular}} & Qwen3-8B & \textbf{\boldmath$-$.242}$^{*}$ & $-$.161 & $-$.022 & \textbf{\boldmath$-$.104}$^{*}$ & .02 \\
 & Gemma-2-9B & $-$.112 & $-$.124 & $-$.047 & \textbf{\boldmath$-$.110}$^{*}$ & .00 \\
 & Mistral-7B & $-$.117 & $-$.034 & .001 & $-$.038 & .04 \\
 & Llama-3.1-8B & .205 & $-$.129 & $-$.024 & $-$.066 & .05 \\
 & Qwen2.5-7B & .000 & $-$.149 & $-$.020 & \textbf{\boldmath$-$.103}$^{*}$ & .02 \\
 & OLMo-3-7B & $-$.099 & $-$.027 & $-$.037 & $-$.028 & .06 \\
\midrule
\multirow{4}{*}{\begin{tabular}[c]{@{}l@{}}Large,\\direct\\answer\end{tabular}} & Llama-3.1-70B & $-$.118 & \textbf{\boldmath$-$.229}$^{*}$ & \textbf{\boldmath$-$.135}$^{*}$ & \textbf{\boldmath$-$.182}$^{*}$ & $-$.14 \\
 & Qwen2.5-72B & .037 & $-$.120 & \textbf{\boldmath$-$.139}$^{*}$ & \textbf{\boldmath$-$.090}$^{*}$ & $-$.10 \\
 & Gemini-2.5-Flash & .014 & $-$.018 & $-$.021 & $-$.018 & $-$.03 \\
 & GPT-4.1 & $-$.109 & $-$.077 & \textbf{\boldmath$-$.104}$^{*}$ & \textbf{\boldmath$-$.093}$^{*}$ & $-$.12 \\
\midrule
\multirow{2}{*}{Reasoning} & Gemini-3.8-Flash & $-$.093 & \textbf{\boldmath$-$.185}$^{*}$ & $-$.022 & $-$.020 & .02 \\
 & Qwen3.8-Flash & $-$.047 & $-$.021 & \textbf{\boldmath$-$.106}$^{*}$ & \textbf{\boldmath$-$.103}$^{*}$ & \textbf{\boldmath$-$.16}$^{*}$ \\
\bottomrule
\end{tabular*}
\caption{\label{tab:feat-interaction}
  Feature-describing generics: standardised prevalence $\times$ cue validity interaction in acceptability,
  with the rater's own prevalence and cue validity (own), the human ones (human P, C; Ryu et al.'s items
  only), or those averaged over the model's group (group avg.\ P, C). Ordinal: the own-ratings model as an
  ordered logit, which allows for the ceiling of the rating scale (log-odds). Bold: a significant negative
  interaction, the human pattern. Ryu et al.: 64 items; ConGen: 582. $^{*}$\,$p < .05$.
}
\end{table*}

\section{Additional factors}
\subsection{What the models' explanations distinguish}
\label{app:cot}

Each 7--9B model rated Ryu et al.’s 84 items after first explaining its reasoning under three chain-of-thought prompts, which gave 1,510 explanations with a valid rating. To see what the models attend to without deciding in advance what to look for, we split the explanations by the rating they ended in, rejection (rating $\leq 0$) or acceptance ($\geq 2$), and asked which words are much more common in one group. We counted every word and two-word phrase, after removing the final answer line, the words of the generic itself, common function words, and words used in fewer than 15 explanations. We scored each by its weighted log-odds ratio with an informative prior \citep{monroe-etal-2008-fightin}, which measures how reliably it distinguishes the two groups.

Table~\ref{tab:cot-words} lists the most distinctive words. Rejections are marked by two kinds of objection: that the statement overgeneralises (\emph{not, all, implies all, generalization, broad, misleading}) and that it is factually wrong (\emph{incorrect, inaccurate, not accurate}). Acceptances are marked by the statement being factually correct (\emph{correct, accurately, fact, known, widely accepted}) and clearly worded (\emph{grammatically correct, clear}), and, less strongly, by the property being \emph{common} or \emph{characteristic}.

\begin{table}
\centering
\small
\begin{tabular}{lr@{\hspace{2em}}lr}
\toprule
\multicolumn{2}{c@{\hspace{2em}}}{\textbf{Rejecting}} & \multicolumn{2}{c}{\textbf{Accepting}} \\
\cmidrule(r{2em}){1-2} \cmidrule{3-4}
Word & $z$ & Word & $z$ \\
\midrule
\emph{not} & 20.1 & \emph{clear} & 10.7 \\
\emph{all} & 14.8 & \emph{correct} & 9.6 \\
\emph{incorrect} & 11.5 & \emph{known} & 8.3 \\
\emph{specific} & 9.7 & \emph{accurately} & 8.3 \\
\emph{implies} & 9.2 & \emph{verb} & 8.0 \\
\emph{generalization} & 8.8 & \emph{widely} & 8.0 \\
\emph{not accurate} & 8.6 & \emph{subject} & 7.7 \\
\emph{misleading} & 8.1 & \emph{grammatically correct} & 7.5 \\
\emph{broad} & 8.0 & \emph{fact} & 7.5 \\
\emph{all not} & 7.9 & \emph{simple} & 7.3 \\
\emph{colors} & 7.5 & \emph{accepted} & 7.1 \\
\emph{inaccurate} & 7.5 & \emph{grammatically} & 7.0 \\
\emph{weighing} & 7.3 & \emph{generally} & 6.7 \\
\emph{diversity} & 7.0 & \emph{common} & 6.6 \\
\emph{general} & 7.0 & \emph{ambiguity} & 6.5 \\
\emph{not entirely} & 6.7 & \emph{characteristic} & 6.2 \\
\emph{hair} & 6.6 & \emph{biological} & 6.0 \\
\emph{specific type} & 6.5 & \emph{no} & 6.0 \\
\emph{implies all} & 6.5 & \emph{conveys} & 5.9 \\
\emph{entirely} & 6.4 & \emph{accurately describes} & 5.9 \\
\bottomrule
\end{tabular}
\caption{\label{tab:cot-words}
  The words and word pairs that most distinguish the models' explanations ending in rejection
  (rating $\leq 0$, $n = 549$) from those ending in acceptance ($\geq 2$, $n = 834$), pooled over
  the six 7--9B models and three prompts: weighted log-odds with an informative prior
  \citep{monroe-etal-2008-fightin}, as $z$-scores; higher means more reliably distinctive. The
  final answer line, the words of the generic itself, common function words (except \emph{not},
  \emph{all}, \emph{no}) and words used in fewer than 15 explanations were removed before counting.}
\end{table}

\subsection{Choice of baseline}
\label{app:baselines}

Table~\ref{tab:baselines} compares three baselines within each type of generic by how well they predict held-out items (10-fold cross-validated $R^2$, mean of five splits): prevalence alone, prevalence and cue validity, and Ryu et al.'s rule with their interaction.

For IS\_A generics, prevalence and cue validity are the best or tied-best baseline in 9 of 12 models on both datasets. On Ryu et al.'s items cue validity takes a negative weight in every model, as it does for people, whose ratings it also predicts better (.71 vs.\ .82): inverses such as \emph{Mammals are dogs} are highly diagnostic but rarely true. Ryu et al.\ describe IS\_A acceptability as following prevalence alone, on the basis of each factor's correlation with acceptability, but did not fit the two together.

For feature-describing generics, prevalence and cue validity are the best baseline, or within .01 of it, in 8 of the 12 models on Ryu et al.’s items and in at least 10 on ConGen. Adding their interaction improves the predictions for only one model on Ryu et al.'s items (Qwen3-8B) and three on ConGen, by at most .02.

\begin{table*}
\centering
\small
\begin{tabular*}{\textwidth}{@{\extracolsep{\fill}}llrr@{\hspace{1.5em}}rr@{\hspace{3em}}rrr@{\hspace{1.5em}}rrr@{}}
\toprule
 & & \multicolumn{4}{c@{\hspace{3em}}}{\textbf{IS\_A}} & \multicolumn{6}{c@{}}{\textbf{Feature-describing}} \\
\cmidrule(r{3em}){3-6} \cmidrule{7-12}
 & & \multicolumn{2}{c@{\hspace{1.5em}}}{Ryu et al.} & \multicolumn{2}{c@{\hspace{3em}}}{ConGen}
 & \multicolumn{3}{c@{\hspace{1.5em}}}{Ryu et al.} & \multicolumn{3}{c@{}}{ConGen} \\
\cmidrule(r{1.5em}){3-4} \cmidrule(r{3em}){5-6} \cmidrule(r{1.5em}){7-9} \cmidrule{10-12}
\textbf{Group} & \textbf{Rater} & P & P+C & P & P+C & P & P+C & P$\times$C & P & P+C & P$\times$C \\
\midrule
Humans & \citet{ryu-etal-2022-flexible} & .71 & \textbf{.82} & -- & -- & .34 & .50 & \textbf{.53} & -- & -- & -- \\
\midrule
\multirow{6}{*}{\begin{tabular}[c]{@{}l@{}}Open,\\7--9B\end{tabular}} & Qwen2.5-7B & \textbf{.26} & .23 & .22 & \textbf{.23} & .16 & \textbf{.23} & .14 & .23 & \textbf{.26} & .25 \\
 & Mistral-7B & \textbf{.54} & .48 & \textbf{.11} & \textbf{.11} & \textbf{.12} & \textbf{.12} & .07 & .24 & \textbf{.26} & \textbf{.26} \\
 & OLMo-3-7B & \textbf{.27} & $-$.84 & .12 & \textbf{.13} & \textbf{\boldmath$-$.01} & $-$.04 & $-$.06 & .22 & \textbf{.23} & \textbf{.23} \\
 & Gemma-2-9B & .66 & \textbf{.79} & .33 & \textbf{.36} & .04 & \textbf{.31} & .29 & .28 & \textbf{.32} & \textbf{.32} \\
 & Qwen3-8B & .66 & \textbf{.81} & .24 & \textbf{.25} & .07 & .14 & \textbf{.16} & .14 & \textbf{.17} & \textbf{.17} \\
 & Llama-3.1-8B & \textbf{.60} & \textbf{.60} & .16 & \textbf{.17} & \textbf{.12} & .05 & $-$.08 & \textbf{.10} & \textbf{.10} & \textbf{.10} \\
\midrule
\multirow{4}{*}{\begin{tabular}[c]{@{}l@{}}Large,\\direct\\answer\end{tabular}} & Llama-3.1-70B & .59 & \textbf{.66} & \textbf{.22} & .20 & .15 & \textbf{.34} & .31 & .12 & .16 & \textbf{.17} \\
 & Qwen2.5-72B & .48 & \textbf{.53} & .34 & \textbf{.35} & .07 & \textbf{.16} & .13 & .32 & .38 & \textbf{.40} \\
 & GPT-4.1 & .47 & \textbf{.90} & \textbf{.30} & \textbf{.30} & $-$.04 & \textbf{.07} & .04 & .13 & .15 & \textbf{.16} \\
 & Gemini-2.5-Flash & .48 & \textbf{.97} & \textbf{.34} & .33 & \textbf{\boldmath$-$.04} & $-$.07 & $-$.11 & .16 & \textbf{.23} & .22 \\
\midrule
\multirow{2}{*}{Reasoning} & Gemini-3.8-Flash & .55 & \textbf{.83} & \textbf{.10} & .09 & $-$.06 & \textbf{.03} & .00 & \textbf{.03} & \textbf{.03} & \textbf{.03} \\
 & Qwen3.8-Flash & .42 & \textbf{.67} & \textbf{.27} & \textbf{.27} & $-$.05 & \textbf{.02} & $-$.01 & \textbf{.08} & \textbf{.08} & \textbf{.08} \\
\bottomrule
\end{tabular*}
\caption{\label{tab:baselines}
  Baselines within each generic type: variance explained in held-out items (10-fold
  cross-validated $R^2$, mean of five splits) by prevalence alone (P), prevalence and cue validity
  (P+C), and Ryu et al.'s rule with their interaction (P$\times$C). Each rater's own ratings.
  Bold: the best baseline (ties included). Negative values: worse than predicting the mean.
  Ryu et al.: 20 IS\_A and 64 feature-describing items; ConGen: 150 and 582.}
\end{table*}

\subsection{Characteristicness and strikingness against prevalence}
\label{app:psych-overlap}

\begin{figure*}[p]
\centering
\includegraphics[width=\textwidth]{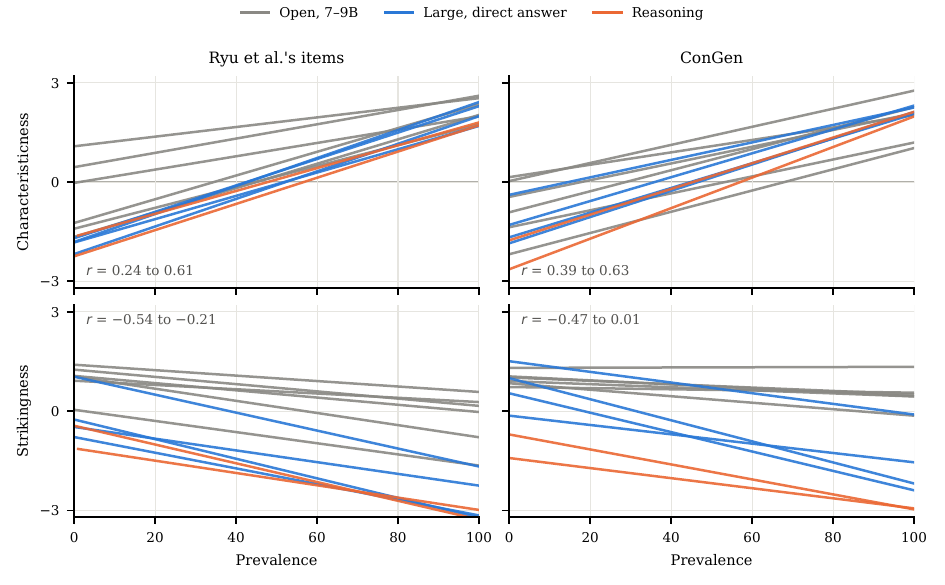}
\caption{\label{fig:psych-overlap}
  Characteristicness and strikingness against prevalence, each model's own ratings: one fitted
  line per model over the range of its prevalence ratings, coloured by group, with the range of
  correlations in each panel. Ratings on $-3$ to $+3$; prevalence on 0--100.
}
\end{figure*}

Figure~\ref{fig:psych-overlap} shows how each model's characteristicness and strikingness ratings relate to its prevalence ratings. Characteristicness rises with prevalence in every model and dataset ($r = .24$--$.63$). This is expected as a principled connection between a kind and a property leads one to expect most members to have it, though a common property need not be a principled one \citep{prasada-etal-2013-conceptual}. Strikingness falls with prevalence in all but one case ($r$ down to $-.54$; the exception is Llama-3.1-8B on ConGen), since striking properties are typically had by few members of a kind \citep{leslie-2008-generics}. The figure also shows a difference between the groups as the 7--9B models rate most generics as mildly striking whatever their prevalence, whereas the larger and reasoning models rate common properties as clearly unremarkable.

\begin{table*}
\centering
\small
\begin{tabular*}{\textwidth}{@{\extracolsep{\fill}}llrrrr@{\hspace{3em}}r@{}}
\toprule
 & & \multicolumn{4}{c@{\hspace{3em}}}{\textbf{Characteristicness}} & \multicolumn{1}{c@{}}{\textbf{Strikingness}} \\
\cmidrule(r{3em}){3-6} \cmidrule{7-7}
 & & \multicolumn{2}{c}{min.\ char.\ $-$ maj.\ stat.} & principled & $\rho$ with & striking \\
\cmidrule{3-4}
\textbf{Group} & \textbf{Model} & raw & given P & $-$ other & humans & $-$ other \\
\midrule
\multirow{6}{*}{\begin{tabular}[c]{@{}l@{}}Open,\\7--9B\end{tabular}} & Qwen2.5-7B & .83$^{*}$ & \textbf{.81}$^{*}$ & 2.44$^{*}$ & .68 & 2.03$^{*}$ \\
 & Mistral-7B & 2.00$^{*}$ & \textbf{1.97}$^{*}$ & 2.41$^{*}$ & .68 & 3.67$^{*}$ \\
 & OLMo-3-7B & $-$.17 & $-$.23 & 1.02$^{*}$ & .68 & 2.28$^{*}$ \\
 & Gemma-2-9B & .25 & .24 & 2.45$^{*}$ & .93$^{*}$ & 4.18$^{*}$ \\
 & Qwen3-8B & 2.50$^{*}$ & \textbf{2.16}$^{*}$ & 1.94$^{*}$ & .85$^{*}$ & 2.36$^{*}$ \\
 & Llama-3.1-8B & .50 & $-$.17 & 1.12$^{*}$ & .85$^{*}$ & 2.39$^{*}$ \\
\midrule
\multirow{4}{*}{\begin{tabular}[c]{@{}l@{}}Large,\\direct\\answer\end{tabular}} & Llama-3.1-70B & $-$1.17 & $-$1.12 & 2.03$^{*}$ & .82$^{*}$ & 5.19$^{*}$ \\
 & Qwen2.5-72B & 1.50$^{*}$ & \textbf{1.41}$^{*}$ & 2.94$^{*}$ & .82$^{*}$ & 4.25$^{*}$ \\
 & GPT-4.1 & 1.45$^{*}$ & \textbf{1.49}$^{*}$ & 3.76$^{*}$ & .93$^{*}$ & 3.94$^{*}$ \\
 & Gemini-2.5-Flash & .75 & 1.24 & 4.00$^{*}$ & .95$^{*}$ & 5.33$^{*}$ \\
\midrule
\multirow{2}{*}{Reasoning} & Gemini-3.8-Flash & 3.42$^{*}$ & \textbf{3.41}$^{*}$ & 4.97$^{*}$ & .95$^{*}$ & 5.32$^{*}$ \\
 & Qwen3.8-Flash & 2.83$^{*}$ & \textbf{3.10}$^{*}$ & 4.20$^{*}$ & .88$^{*}$ & 4.17$^{*}$ \\
\bottomrule
\end{tabular*}
\caption{\label{tab:validate-factors}
  Validation of the characteristicness and strikingness ratings on Ryu et al.'s items, which
  are the materials of \citet{prasada-etal-2013-conceptual}, twelve of each of seven types.
  Min.\ char.\ $-$ maj.\ stat.: mean characteristicness of minority characteristic generics
  (\emph{Ducks lay eggs}) minus majority statistical ones (\emph{Cars have radios}), raw and
  controlling for the model's own prevalence (in Ryu et al.'s human ratings the former are the less
  prevalent, 69\% vs.\ 74\%). Principled $-$ other: quasi-definitions and majority and minority
  characteristic generics minus the other four types. $\rho$: rank correlation of the model's seven
  type means with Prasada et al.'s human ratings of principled connections. Striking $-$ other: mean
  strikingness of the striking generics minus all others. Ratings on $-3$ to $+3$. Bold: the contrast
  holds with prevalence controlled. $^{*}$\,$p < .05$.}
\end{table*}

\subsection{Validation of characteristicness and strikingness}
\label{app:validate}

Ryu et al.'s 84 items are the materials of \citet{prasada-etal-2013-conceptual}, who sorted them into
seven types of twelve items each. Their account predicts which types involve a principled connection between kind and property, and so should be rated characteristic: quasi-definitions (\emph{Ants are insects}) and majority and minority characteristic generics
(\emph{Tigers are striped}, \emph{Ducks lay eggs}). The other four types should not be: majority statistical generics (\emph{Cars have radios}), striking generics (\emph{Mosquitoes carry malaria}),
and majority and minority false generalisations (\emph{Canadians are right-handed}, \emph{Rooms are round}). Table~\ref{tab:validate-factors} tests each model's ratings against these predictions. All twelve models rate the three principled types as more characteristic than the other four, and for nine of them the ranking of the seven types follows people's ratings of principled connections in Prasada et al.'s study.

The hardest case contrasts minority characteristic generics such as \emph{Ducks lay eggs} with majority statistical ones such as \emph{Cars have radios}. Elsewhere, characteristic properties are also the more common ones, so a model could pass by tracking prevalence; here prevalence points the other way, as people judge the former less common, so a model relying on prevalence would rate them lower. We therefore also compare the two types while controlling for each model's own prevalence ratings. Seven of the twelve models rate minority characteristic generics as more characteristic even then. In the other five the difference is not significant: it is near zero or reverses for OLMo-3-7B, Gemma-2-9B and the two Llama models, and is positive but not significant for Gemini-2.5-Flash ($p = .055$). For Llama-3.1-8B, the raw difference disappears once prevalence is controlled, so its characteristicness ratings partly re-express its prevalence estimates. Several of these models also rate striking properties as characteristic, which \citet{prasada-etal-2013-conceptual} do not count as principled. With twelve items per type, these tests have limited power, but they suggest that the characteristicness weights in Section~\ref{sec:factors} are most trustworthy for the seven models that pass.

\subsection{Full results for the additional factors}
\label{app:factor-tables}\label{app:psych-isa}\label{app:model-factors-isa}

Table~\ref{tab:psych-factors} adds characteristicness and strikingness to the baseline, and Table~\ref{tab:model-factors} adds factuality and overreach to all four psychological factors, for feature-describing generics on both datasets and IS\_A generics on ConGen. Together they give, for every model, the variance explained at each step summarised in Figure~\ref{fig:factor-summary}. Characteristicness raises acceptability in most models (eleven of twelve on Ryu et al.'s feature-describing items, nine on ConGen's, eleven on ConGen's IS\_A items), while strikingness is mostly near zero and, where significant, more often negative. Factuality has a significant positive weight in every model and dataset; overreach, where significant, mostly lowers acceptability. Most of what the four psychological factors explain in acceptability is also explained by factuality and overreach.

Ryu et al.'s 20 IS\_A items are left out of both tables. Across so few items, prevalence, cue validity and characteristicness are so closely related that each can largely be predicted from the others (variance inflation factors up to 60, against at most 2.2 elsewhere), and the four psychological factors explain 75--99\% of the variance in factuality, so the weights are not interpretable. Strikingness also cannot be estimated for four models, which rate every IS\_A item as equally unremarkable.

\begin{table*}
\centering
\small
\setlength{\tabcolsep}{3pt}
\begin{tabular*}{\textwidth}{@{\extracolsep{\fill}}llrrrr@{\hspace{1.5em}}rrrr@{\hspace{1.5em}}rrrr@{}}
\toprule
 & & \multicolumn{4}{c@{\hspace{1.5em}}}{\textbf{Feature, Ryu et al.}} & \multicolumn{4}{c@{\hspace{1.5em}}}{\textbf{Feature, ConGen}} & \multicolumn{4}{c@{}}{\textbf{IS\_A, ConGen}} \\
\cmidrule(r{1.5em}){3-6} \cmidrule(r{1.5em}){7-10} \cmidrule{11-14}
\textbf{Group} & \textbf{Model} & $R^2_{P,C}$ & Ch & S & $R^2$ & $R^2_{P,C}$ & Ch & S & $R^2$ & $R^2_{P,C}$ & Ch & S & $R^2$ \\
\midrule
\multirow{6}{*}{\begin{tabular}[c]{@{}l@{}}Open,\\7--9B\end{tabular}}
 & Qwen2.5-7B & .36 & .35$^{*}$ & .26$^{*}$ & \textbf{.47}$^{*}$ & .27 & .09$^{*}$ & $-$.04 & \textbf{.28}$^{*}$ & .27 & .28$^{*}$ & $-$.01 & \textbf{.32}$^{*}$ \\
 & Mistral-7B & .23 & .32$^{*}$ & .17 & \textbf{.33}$^{*}$ & .27 & .22$^{*}$ & .02 & \textbf{.31}$^{*}$ & .14 & .27$^{*}$ & $-$.08 & \textbf{.22}$^{*}$ \\
 & OLMo-3-7B & .05 & .50$^{*}$ & .02 & \textbf{.26}$^{*}$ & .24 & .06 & $-$.06 & .25 & .16 & .22$^{*}$ & .06 & \textbf{.20}$^{*}$ \\
 & Gemma-2-9B & .37 & .39$^{*}$ & .09 & \textbf{.51}$^{*}$ & .33 & .17$^{*}$ & $-$.01 & \textbf{.35}$^{*}$ & .38 & .33$^{*}$ & .01 & \textbf{.45}$^{*}$ \\
 & Qwen3-8B & .21 & .18 & $-$.10 & .24 & .18 & .13$^{*}$ & .05 & \textbf{.20}$^{*}$ & .28 & .12 & $-$.09 & .30 \\
 & Llama-3.1-8B & .22 & .47$^{*}$ & $-$.12 & \textbf{.37}$^{*}$ & .12 & .09$^{*}$ & .08$^{*}$ & \textbf{.13}$^{*}$ & .21 & .25$^{*}$ & $-$.01 & \textbf{.26}$^{*}$ \\
\midrule
\multirow{4}{*}{\begin{tabular}[c]{@{}l@{}}Large,\\direct\\answer\end{tabular}}
 & Llama-3.1-70B & .42 & .26$^{*}$ & .14 & \textbf{.49}$^{*}$ & .17 & .06 & .01 & .17 & .25 & .31$^{*}$ & $-$.05 & \textbf{.33}$^{*}$ \\
 & Qwen2.5-72B & .24 & .59$^{*}$ & .09 & \textbf{.47}$^{*}$ & .39 & .22$^{*}$ & .02 & \textbf{.42}$^{*}$ & .38 & .19$^{*}$ & $-$.09 & \textbf{.41}$^{*}$ \\
 & GPT-4.1 & .18 & .61$^{*}$ & $-$.03 & \textbf{.44}$^{*}$ & .16 & .13$^{*}$ & $-$.18$^{*}$ & \textbf{.20}$^{*}$ & .33 & .22$^{*}$ & $-$.10 & \textbf{.38}$^{*}$ \\
 & Gemini-2.5-Flash & .03 & .76$^{*}$ & $-$.35$^{*}$ & \textbf{.50}$^{*}$ & .23 & .18$^{*}$ & $-$.07 & \textbf{.26}$^{*}$ & .35 & .32$^{*}$ & $-$.21$^{*}$ & \textbf{.44}$^{*}$ \\
\midrule
\multirow{2}{*}{Reasoning}
 & Gemini-3.8-Flash & .10 & .59$^{*}$ & .03 & \textbf{.36}$^{*}$ & .05 & .10$^{*}$ & $-$.14$^{*}$ & \textbf{.07}$^{*}$ & .14 & .24$^{*}$ & $-$.17$^{*}$ & \textbf{.20}$^{*}$ \\
 & Qwen3.8-Flash & .12 & .50$^{*}$ & .05 & \textbf{.33}$^{*}$ & .09 & .00 & $-$.05 & .09 & .31 & .24$^{*}$ & $-$.02 & \textbf{.34}$^{*}$ \\
\bottomrule
\end{tabular*}
\caption{\label{tab:psych-factors}\label{tab:psych-factors-feature}\label{tab:psych-factors-is_a}
  Characteristicness (Ch) and strikingness (S) added to the baseline of prevalence and cue validity:
  standardised weights in each model's acceptability, each model's own ratings. $R^2_{P,C}$: variance
  explained by the baseline; $R^2$: with Ch and S added, bold when the rise is significant ($F$ test).
  Ryu et al.: 64 items; ConGen: 582 feature-describing and 150 IS\_A. $^{*}$\,$p < .05$.}
\end{table*}

\begin{table*}
\centering
\small
\setlength{\tabcolsep}{3pt}
\begin{tabular*}{\textwidth}{@{\extracolsep{\fill}}llrrrr@{\hspace{1.5em}}rrrr@{\hspace{1.5em}}rrrr@{}}
\toprule
 & & \multicolumn{4}{c@{\hspace{1.5em}}}{\textbf{Feature, Ryu et al.}} & \multicolumn{4}{c@{\hspace{1.5em}}}{\textbf{Feature, ConGen}} & \multicolumn{4}{c@{}}{\textbf{IS\_A, ConGen}} \\
\cmidrule(r{1.5em}){3-6} \cmidrule(r{1.5em}){7-10} \cmidrule{11-14}
\textbf{Group} & \textbf{Model} & F & O & $R^2_6$ & shared & F & O & $R^2_6$ & shared & F & O & $R^2_6$ & shared \\
\midrule
\multirow{6}{*}{\begin{tabular}[c]{@{}l@{}}Open,\\7--9B\end{tabular}}
 & Qwen2.5-7B & .58$^{*}$ & .19$^{*}$ & .68$^{*}$ & 90\% & .72$^{*}$ & .01 & .64$^{*}$ & 96\% & .72$^{*}$ & .06 & .66$^{*}$ & 96\% \\
 & Mistral-7B & .59$^{*}$ & \textbf{\boldmath$-$.28}$^{*}$ & .64$^{*}$ & 98\% & .58$^{*}$ & \textbf{\boldmath$-$.12}$^{*}$ & .58$^{*}$ & 91\% & .65$^{*}$ & $-$.11 & .56$^{*}$ & 97\% \\
 & OLMo-3-7B & .35$^{*}$ & $-$.16 & .42$^{*}$ & 43\% & .57$^{*}$ & .03 & .49$^{*}$ & 85\% & .77$^{*}$ & \textbf{\boldmath$-$.13}$^{*}$ & .71$^{*}$ & 96\% \\
 & Gemma-2-9B & .42$^{*}$ & $-$.22 & .67$^{*}$ & 89\% & .56$^{*}$ & \textbf{\boldmath$-$.20}$^{*}$ & .63$^{*}$ & 97\% & .59$^{*}$ & \textbf{\boldmath$-$.30}$^{*}$ & .80$^{*}$ & 99\% \\
 & Qwen3-8B & .59$^{*}$ & \textbf{\boldmath$-$.25}$^{*}$ & .57$^{*}$ & 86\% & .56$^{*}$ & \textbf{\boldmath$-$.17}$^{*}$ & .51$^{*}$ & 94\% & .70$^{*}$ & $-$.06 & .70$^{*}$ & 89\% \\
 & Llama-3.1-8B & .63$^{*}$ & $-$.05 & .64$^{*}$ & 80\% & .42$^{*}$ & .08$^{*}$ & .29$^{*}$ & 62\% & .28$^{*}$ & .09 & .32$^{*}$ & 53\% \\
\midrule
\multirow{4}{*}{\begin{tabular}[c]{@{}l@{}}Large,\\direct\\answer\end{tabular}}
 & Llama-3.1-70B & .55$^{*}$ & .00 & .63$^{*}$ & 89\% & .56$^{*}$ & $-$.05 & .44$^{*}$ & 91\% & .53$^{*}$ & $-$.07 & .54$^{*}$ & 88\% \\
 & Qwen2.5-72B & .83$^{*}$ & $-$.03 & .85$^{*}$ & 95\% & .65$^{*}$ & \textbf{\boldmath$-$.12}$^{*}$ & .68$^{*}$ & 96\% & .70$^{*}$ & $-$.10 & .71$^{*}$ & 98\% \\
 & GPT-4.1 & .71$^{*}$ & $-$.06 & .71$^{*}$ & 81\% & .62$^{*}$ & \textbf{\boldmath$-$.10}$^{*}$ & .49$^{*}$ & 97\% & .60$^{*}$ & \textbf{\boldmath$-$.21}$^{*}$ & .64$^{*}$ & 98\% \\
 & Gemini-2.5-Flash & .34$^{*}$ & $-$.09 & .57$^{*}$ & 54\% & .48$^{*}$ & \textbf{\boldmath$-$.12}$^{*}$ & .44$^{*}$ & 92\% & .53$^{*}$ & \textbf{\boldmath$-$.17}$^{*}$ & .64$^{*}$ & 96\% \\
\midrule
\multirow{2}{*}{Reasoning}
 & Gemini-3.8-Flash & .49$^{*}$ & $-$.20 & .54$^{*}$ & 73\% & .21$^{*}$ & $-$.01 & .10$^{*}$ & 74\% & .27$^{*}$ & $-$.10 & .26$^{*}$ & 86\% \\
 & Qwen3.8-Flash & .64$^{*}$ & $-$.10 & .49$^{*}$ & 73\% & .35$^{*}$ & \textbf{\boldmath$-$.13}$^{*}$ & .20$^{*}$ & 88\% & .42$^{*}$ & \textbf{\boldmath$-$.23}$^{*}$ & .49$^{*}$ & 94\% \\
\bottomrule
\end{tabular*}
\caption{\label{tab:model-factors}\label{tab:model-factors-feature}\label{tab:model-factors-is_a}
  Factuality (F) and overreach (O) added to the four psychological factors (prevalence, cue validity,
  characteristicness, strikingness): standardised weights, each model's own ratings. Factuality is the
  largest of the six weights in every cell except OLMo-3-7B (characteristicness) and Gemini-2.5-Flash
  (prevalence) on Ryu et al.'s items. Bold: overreach significantly lowers acceptability. $R^2_6$:
  variance explained by all six factors, starred when the rise over the four (the $R^2$ of
  Table~\ref{tab:psych-factors}) is significant ($F$ test). Shared: of what the four factors explain
  in acceptability, the share that factuality and overreach also explain. $^{*}$\,$p < .05$.}
\end{table*}

\subsection{Interactions of the psychological factors}
\label{app:psych-interactions}

The psychological accounts predict two interactions with prevalence. \citet{leslie-2008-generics} proposes that a striking property licenses a generic even when it is rare, so strikingness should matter more at low prevalence (a negative prevalence $\times$ strikingness weight). A characteristic property may likewise compensate for low prevalence, as in \emph{Ducks lay eggs} (a negative prevalence $\times$ characteristicness weight). To test them, we add each interaction, one at a time, to the four-factor regression of Table~\ref{tab:psych-factors} and report its weight in Table~\ref{tab:psych-interactions}, together with the change in held-out $R^2$ when all pairwise interactions are added at once.

Neither prediction holds reliably. No model shows the first: where the prevalence $\times$ strikingness weight is significant, it is positive. The second appears in three models for feature-describing generics on ConGen, with small weights, and in the opposite direction in two on Ryu et al.'s items. Adding all pairwise interactions among the factors at once makes the predictions for held-out items worse in 11 of 12 models for feature-describing generics on both datasets and in all 12 for IS\_A generics on ConGen. We therefore report the additive models.

\begin{table*}
\centering
\small
\setlength{\tabcolsep}{3pt}
\begin{tabular*}{\textwidth}{@{\extracolsep{\fill}}llrrr@{\hspace{1.5em}}rrr@{\hspace{1.5em}}rrr@{}}
\toprule
 & & \multicolumn{3}{c@{\hspace{1.5em}}}{\textbf{Feature, Ryu et al.}} & \multicolumn{3}{c@{\hspace{1.5em}}}{\textbf{Feature, ConGen}} & \multicolumn{3}{c@{}}{\textbf{IS\_A, ConGen}} \\
\cmidrule(r{1.5em}){3-5} \cmidrule(r{1.5em}){6-8} \cmidrule{9-11}
\textbf{Group} & \textbf{Model} & P$\times$Ch & P$\times$S & $\Delta R^2_{\mathrm{cv}}$ & P$\times$Ch & P$\times$S & $\Delta R^2_{\mathrm{cv}}$ & P$\times$Ch & P$\times$S & $\Delta R^2_{\mathrm{cv}}$ \\
\midrule
\multirow{6}{*}{\begin{tabular}[c]{@{}l@{}}Open,\\7--9B\end{tabular}} & Qwen2.5-7B & $-$.09 & $-$.02 & $-$.074 & .00 & $-$.06 & $-$.011 & $-$.07 & .00 & $-$.073 \\
 & Mistral-7B & .01 & $-$.23 & $-$.154 & $-$.04 & .00 & $-$.019 & $-$.01 & .02 & $-$.115 \\
 & OLMo-3-7B & $-$.12 & .25$^{*}$ & $-$.154 & $-$.01 & .09$^{*}$ & $-$.007 & $-$.17 & .01 & $-$.051 \\
 & Gemma-2-9B & .04 & $-$.16 & $-$.428 & $-$.01 & $-$.02 & $-$.018 & .00 & $-$.11 & $-$.046 \\
 & Qwen3-8B & $-$.09 & .13 & $-$.170 & \textbf{\boldmath$-$.12}$^{*}$ & $-$.02 & $-$.016 & $-$.04 & $-$.10 & $-$.064 \\
 & Llama-3.1-8B & .16 & .12 & $-$.622 & $-$.07 & .00 & $-$.020 & $-$.08 & .05 & $-$.079 \\
\midrule
\multirow{4}{*}{\begin{tabular}[c]{@{}l@{}}Large,\\direct\\answer\end{tabular}} & Llama-3.1-70B & .13 & .03 & .014 & \textbf{\boldmath$-$.10}$^{*}$ & .07$^{*}$ & $-$.001 & $-$.01 & .00 & $-$.193 \\
 & Qwen2.5-72B & .17 & $-$.11 & $-$.238 & \textbf{\boldmath$-$.06}$^{*}$ & $-$.04 & .000 & $-$.07 & .11 & $-$.028 \\
 & GPT-4.1 & .19 & .16 & $-$.024 & $-$.04 & .06 & $-$.021 & .10 & $-$.08 & $-$.098 \\
 & Gemini-2.5-Flash & .22$^{*}$ & $-$.10 & $-$1.154 & .01 & .02 & $-$.008 & .11 & $-$.02 & $-$.098 \\
\midrule
\multirow{2}{*}{Reasoning} & Gemini-3.8-Flash & .18 & .08 & $-$.049 & $-$.07 & .09$^{*}$ & $-$.024 & .17 & .00 & $-$.147 \\
 & Qwen3.8-Flash & .25$^{*}$ & $-$.01 & $-$.082 & $-$.01 & $-$.01 & $-$.013 & .13 & $-$.01 & $-$.116 \\
\bottomrule
\end{tabular*}
\caption{\label{tab:psych-interactions}\label{tab:psych-interactions-feature}\label{tab:psych-interactions-is_a}
  Interactions predicted by the psychological accounts, each added on its own to the model of
  Table~\ref{tab:psych-factors}: prevalence $\times$ characteristicness (P$\times$Ch) and
  prevalence $\times$ strikingness (P$\times$S), standardised weights; both are predicted to be
  negative (bold: significant and negative). $\Delta R^2_{\mathrm{cv}}$: change in held-out
  $R^2$ (10-fold cross-validation) from adding all pairwise interactions among the factors;
  negative: they predict new items worse. $^{*}$\,$p < .05$.}
\end{table*}

\subsection{Shared variance of the two sets of factors}
\label{app:variance-partition}

\begin{figure*}[p]
\centering
\includegraphics[width=\textwidth]{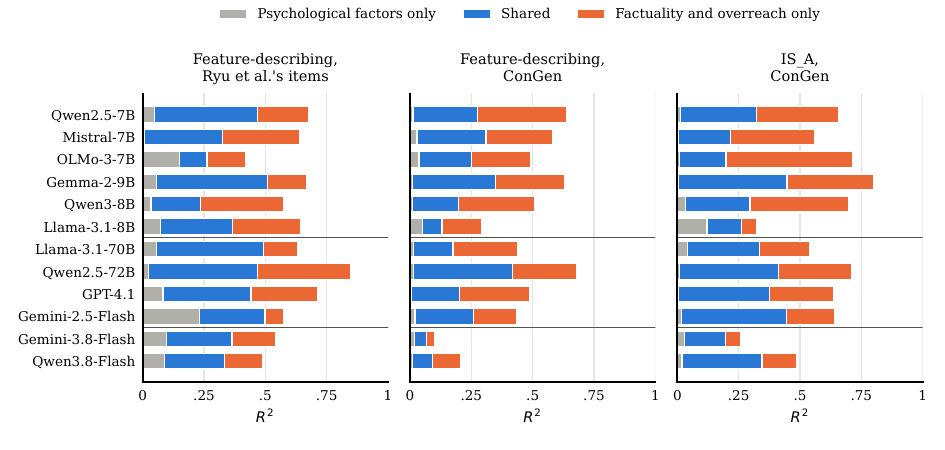}
\caption{\label{fig:variance-partition}
  What the six factors explain in each model's acceptability, split into the part explained only by
  the psychological factors (prevalence, cue validity, characteristicness, strikingness), the part
  explained by either set, and the part explained only by factuality and overreach (in-sample
  $R^2$, each model's own ratings). Ryu et al.'s 20 IS\_A items are too few for this analysis
  (Appendix~\ref{app:model-factors-isa}).
}
\end{figure*}

Figure~\ref{fig:variance-partition} splits what the six factors explain in each model's
acceptability into three parts (a commonality analysis): the part explained only by the
psychological factors, the part explained by either set, and the part explained only by factuality
and overreach. The shared part is $R^2_{\mathit{psych}} + R^2_{F,O} - R^2_{\mathit{all}}$, where each
$R^2$ comes from a model with that set of factors.

The part explained only by the psychological factors is small in almost every model, especially on ConGen: almost everything they explain about acceptability is also explained by factuality and overreach. This is because the models’ factuality and overreach ratings themselves depend on the psychological factors. On ConGen, prevalence and characteristicness predict higher factuality in 11--12 of the 12 models and lower overreach in 9--10, cue validity predicts higher factuality for feature-describing generics in 11, and strikingness predicts neither. On top of this shared part, factuality and overreach explain about as much again of their own. 

\subsection{Surprisal-based acceptability}
\label{app:surprisal}

For the six 7--9B models, whose token probabilities are available, we computed two measures of
acceptability that involve no question. P-Acceptability is the negative cross-entropy of the property
tokens given the subject, so that a property the model finds predictable for the kind scores high
\citep{cilleruelo-etal-2025-generics}. SLOR is the log-probability of the whole sentence, normalised
for the frequency of its words and its length \citep{lau-etal-2017-grammaticality}. On Ryu et al.'s items, P-Acceptability correlates with people's acceptability at $r = .39$--$.46$ and SLOR at $.29$--$.43$, both below the models' answers ($r = .51$--$.84$; Table~\ref{tab:surprisal-signal}).

\paragraph{Ryu et al.'s analysis.} Table~\ref{tab:surprisal-signal} also repeats Ryu et al.'s
analysis with P-Acceptability as the measure of acceptability and people's prevalence and cue
validity as predictors. For IS\_A generics, prevalence raises P-Acceptability in all six models
(significantly in two, with 20 items) and cue validity lowers it in five, as for people. For
feature-describing generics, cue validity raises it in all six (significantly in five), and the
interaction of prevalence and cue validity is negative, the human direction, in all six, but
significant in none. The human pattern thus appears not only in what the models answer but in how probable they find the generic itself: its main effects are clearly present, while the interaction, as with the written ratings, points the right way but is too weak to establish for individual models.

\begin{figure}[t]
\centering
\includegraphics[width=\columnwidth]{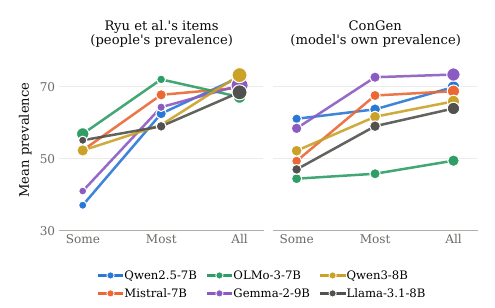}
\caption{\label{fig:quantifier-prevalence}
  Mean prevalence of the generics best fitted by \emph{Some}, \emph{Most} and \emph{All}
  \citep{cilleruelo-etal-2025-generics}, per model: people's prevalence on Ryu et al.'s items, the
  model's own on ConGen. Marker size: share of items with that best fit. The bare plural is not
  shown.}
\end{figure}

\paragraph{Quantifiers.} Following \citet{cilleruelo-etal-2025-generics}, we compared each generic with
its quantified versions and recorded the version (\emph{All}, \emph{Most}, \emph{Some}, or the bare plural) whose property the model finds most predictable. \emph{All} is the most frequent best fit for every model (37--71\% of Ryu et al.'s items, 30--42\% of ConGen's), \emph{Some} accounts for 19--27\% of ConGen's, close to the fifth that \citet{cilleruelo-etal-2025-generics} report, and the bare plural fits best for only 2--24\%. We also find that the best-fitting quantifier tracks prevalence (Figure~\ref{fig:quantifier-prevalence}): in every model and dataset except OLMo on Ryu et al.'s items, the generics best fitted by \emph{All} have the highest prevalence and those best fitted by \emph{Some} the lowest (people's prevalence on Ryu et al.'s items, the model's own on ConGen).

\begin{table*}
\centering
\small
\begin{tabular*}{\textwidth}{@{\extracolsep{\fill}}lrrr@{\hspace{2em}}rr@{\hspace{2em}}rrr@{}}
\toprule
 & \multicolumn{3}{c@{\hspace{2em}}}{$r$ with people's acceptability} & \multicolumn{5}{c@{}}{P-Acceptability from people's P and C} \\
\cmidrule(r{2em}){2-4} \cmidrule{5-9}
 & & & & \multicolumn{2}{c@{\hspace{2em}}}{IS\_A} & \multicolumn{3}{c@{}}{Feature-describing} \\
\cmidrule(r{2em}){5-6} \cmidrule{7-9}
\textbf{Model} & P-Acc. & SLOR & answers & P & C & P & C & P$\times$C \\
\midrule
Qwen2.5-7B & .44$^{*}$ & .43$^{*}$ & .72$^{*}$ & .46 & $-$.16 & .20 & \textbf{.32}$^{*}$ & $-$.07 \\
Mistral-7B & .42$^{*}$ & .37$^{*}$ & .67$^{*}$ & \textbf{.61}$^{*}$ & $-$.08 & .12 & \textbf{.45}$^{*}$ & $-$.03 \\
OLMo-3-7B & .44$^{*}$ & .38$^{*}$ & .51$^{*}$ & .41 & $-$.42 & .23 & \textbf{.30}$^{*}$ & .00 \\
Gemma-2-9B & .46$^{*}$ & .29$^{*}$ & .84$^{*}$ & \textbf{.77}$^{*}$ & .04 & \textbf{.42$^{*}$} & \textbf{.25}$^{*}$ & $-$.09 \\
Qwen3-8B & .39$^{*}$ & .30$^{*}$ & .69$^{*}$ & .21 & $-$.49 & .18 & .23 & $-$.11 \\
Llama-3.1-8B & .44$^{*}$ & .38$^{*}$ & .77$^{*}$ & .51 & $-$.10 & .21 & \textbf{.31}$^{*}$ & $-$.07 \\
\bottomrule
\end{tabular*}
\caption{\label{tab:surprisal-signal}
  Surprisal-based acceptability of the 7--9B models on Ryu et al.'s items. Left: correlation with
  people's acceptability of P-Acceptability \citep{cilleruelo-etal-2025-generics}, of SLOR
  \citep{lau-etal-2017-grammaticality}, and of the model's answers for comparison. Right: Ryu et al.'s
  analysis with P-Acceptability as the acceptability measure and people's prevalence (P) and cue
  validity (C) as predictors, standardised weights; 20 IS\_A and 64 feature-describing items. Bold:
  a significant effect in the human direction. $^{*}$\,$p < .05$.}
\end{table*}

\begin{table*}
\centering
\small
\begin{tabular*}{\textwidth}{@{\extracolsep{\fill}}llrrrrrrr@{}}
\toprule
\textbf{Model} & \textbf{Layer} & F & P & C & Ch & O & S & Random \\
\midrule
Qwen2.5-7B & 15 / 28 & \textbf{.64}$^{*}$ & .50$^{*}$ & .31$^{*}$ & .18$^{*}$ & \textbf{\boldmath$-$.09}$^{*}$ & .21$^{*}$ & $-$.07$^{*}$ \\
Llama-3.1-8B & 13 / 32 & \textbf{1.68}$^{*}$ & 1.27$^{*}$ & 1.57$^{*}$ & 1.44$^{*}$ & \textbf{\boldmath$-$.07}$^{*}$ & .43$^{*}$ & $-$.08$^{*}$ \\
Mistral-7B & 13 / 32 & \textbf{1.99}$^{*}$ & 1.74$^{*}$ & 1.40$^{*}$ & 1.09$^{*}$ & \textbf{\boldmath$-$.87}$^{*}$ & .50$^{*}$ & .05$^{*}$ \\
OLMo-3-7B & 13 / 32 & \textbf{2.44}$^{*}$ & 1.58$^{*}$ & 1.35$^{*}$ & .93$^{*}$ & \textbf{\boldmath$-$.47}$^{*}$ & .38$^{*}$ & .11$^{*}$ \\
Gemma-2-9B & 17 / 42 & \textbf{.67}$^{*}$ & .51$^{*}$ & .57$^{*}$ & .40$^{*}$ & \textbf{\boldmath$-$.59}$^{*}$ & .36$^{*}$ & .01 \\
Qwen3-8B & 20 / 36 & \textbf{1.22}$^{*}$ & 1.11$^{*}$ & .99$^{*}$ & 1.10$^{*}$ & \textbf{\boldmath$-$.60}$^{*}$ & .43$^{*}$ & $-$.10$^{*}$ \\
\bottomrule
\end{tabular*}
\caption{\label{tab:steering-written}
  As Table~\ref{tab:steering}, with each factor's direction fitted to the rating the model wrote
  instead of its expected rating. Bold: the largest positive effect (factuality), and overreach,
  which lowers the rating as its meaning predicts. $^{*}$\,more than two standard errors from zero.}
\end{table*}

\section{Mechanistic analysis}

\subsection{Results with the written ratings}
\label{app:written-labels}

In the main analysis, each factor's direction is fitted to the model's expected rating of the factor, the average of the possible answers weighted by their probabilities. Fitting the directions to the ratings the models actually wrote instead gives the same results. Across all layers and directions, the steering effects of the two versions correlate at $r = .993$--$.998$ for every model, and the layer at which factuality's effect peaks is the same in all six (Table~\ref{tab:steering-written}).

\subsection{What the models write and what their probabilities say}
\label{app:faithfulness}

A model's written rating need not be the answer its output probabilities favor, so we compared the two for every question (Table~\ref{tab:logits-vs-answers}). For each of ConGen's feature-describing generics, the expected rating is the average of the possible answers weighted by their probabilities when the model is about to answer, and the most probable answer is the single answer with the highest probability; the written rating comes from greedy generation at temperature~0.

The two agree closely but not perfectly. The written rating correlates with the expected rating at $r = .77$--$.99$, depending on model and question, and nearly all probability falls on valid answers for every model and question. Agreement differs more between models than between questions: Qwen3-8B writes its most probable answer for 95--99\% of items ($r = .97$--$.99$), whereas OLMo does so for as few as 58\% ($r = .77$--$.91$). Among the questions, overreach shows the least agreement ($r = .77$--$.97$).

Even with greedy decoding, a written rating can differ from the most probable answer for technical reasons: a number is written token by token, so the first token chosen need not begin the most probable number (as for prevalence, from 0 to 100), and the prompts of the written runs and of the scoring differ slightly in tokenization for some models. The disagreement is therefore an upper bound on how far what a model writes departs from what it encodes about the answer. In line with \citet{hu-levy-2023-prompting}, it suggests that written ratings are a close but imperfect window on a model's own judgment, which is why the main analysis uses the expected ratings; as Appendix~\ref{app:written-labels} shows, the conclusions do not depend on the choice.

\begin{table*}
\centering
\small
\setlength{\tabcolsep}{3.5pt}
\begin{tabular*}{\textwidth}{@{\extracolsep{\fill}}lrrrrrrrrrrrrrrr@{}}
\toprule
 & \multicolumn{2}{c}{A} & \multicolumn{2}{c}{P} & \multicolumn{2}{c}{C} & \multicolumn{2}{c}{F} & \multicolumn{2}{c}{Ch} & \multicolumn{2}{c}{O} & \multicolumn{2}{c}{S} & \\
\cmidrule(lr){2-3} \cmidrule(lr){4-5} \cmidrule(lr){6-7} \cmidrule(lr){8-9} \cmidrule(lr){10-11} \cmidrule(lr){12-13} \cmidrule(lr){14-15}
\textbf{Model} & $r$ & =mode & $r$ & =mode & $r$ & =mode & $r$ & =mode & $r$ & =mode & $r$ & =mode & $r$ & =mode & mass \\
\midrule
Qwen2.5-7B & .98 & 90\% & .97 & 80\% & .96 & 86\% & .97 & 86\% & .96 & \textbf{73\%} & .92 & 79\% & .94 & 88\% & 1.00 \\
Mistral-7B & .92 & 87\% & \textbf{.87} & 71\% & .89 & 83\% & .92 & 86\% & .91 & 77\% & .87 & 74\% & .90 & 87\% & 1.00 \\
OLMo-3-7B & \textbf{.87} & \textbf{82\%} & \textbf{.87} & \textbf{58\%} & \textbf{.81} & \textbf{79\%} & \textbf{.86} & \textbf{73\%} & \textbf{.89} & 91\% & \textbf{.77} & 97\% & .91 & 95\% & 1.00 \\
Gemma-2-9B & .97 & 86\% & .95 & 82\% & .97 & 90\% & .97 & 92\% & .96 & 87\% & .94 & 87\% & .95 & \textbf{86\%} & 1.00 \\
Qwen3-8B & .99 & 97\% & .98 & 95\% & .99 & 97\% & .99 & 99\% & .99 & 96\% & .97 & 98\% & .98 & 98\% & 1.00 \\
Llama-3.1-8B & .94 & 90\% & .95 & 96\% & .94 & 91\% & .89 & 95\% & .92 & 93\% & .84 & \textbf{71\%} & \textbf{.87} & 88\% & 1.00 \\
\bottomrule
\end{tabular*}
\caption{\label{tab:logits-vs-answers}
  What the 7--9B models write against what their output probabilities say, ConGen's
  feature-describing generics. $r$: correlation of the written answer with the expected rating
  (the probability-weighted mean of the valid answers); =mode: share of items on which the
  written answer is the single most probable answer; mass: mean probability on valid answers,
  averaged over the questions. Bold: the lowest agreement for each question. A: acceptability,
  P: prevalence, C: cue validity, F: factuality, Ch: characteristicness, O: overreach,
  S: strikingness.}
\end{table*}

\subsection{Swaps}
\label{app:swaps}

\begin{table*}
\centering
\small
\begin{tabular*}{\textwidth}{@{\extracolsep{\fill}}llrrrrrrr@{}}
\toprule
\textbf{Model} & \textbf{Layer} & F & P & C & Ch & O & S & Random \\
\midrule
Qwen2.5-7B & 15 / 28 & \textbf{.41}$^{*}$ & .11$^{*}$ & .04$^{*}$ & .00 & .00 & .04$^{*}$ & .00 \\
Llama-3.1-8B & 13 / 32 & \textbf{1.28}$^{*}$ & .72$^{*}$ & .43$^{*}$ & .38$^{*}$ & \textbf{\boldmath$-$.10}$^{*}$ & .04$^{*}$ & .00 \\
Mistral-7B & 13 / 32 & \textbf{1.79}$^{*}$ & .68$^{*}$ & .58$^{*}$ & .26$^{*}$ & \textbf{\boldmath$-$.58}$^{*}$ & .04$^{*}$ & .00 \\
OLMo-3-7B & 13 / 32 & \textbf{1.37}$^{*}$ & .32$^{*}$ & .18$^{*}$ & .10$^{*}$ & \textbf{\boldmath$-$.13}$^{*}$ & .02$^{*}$ & .01 \\
Gemma-2-9B & 17 / 42 & .05 & \textbf{.11}$^{*}$ & .06$^{*}$ & .04$^{*}$ & \textbf{\boldmath$-$.19}$^{*}$ & .03$^{*}$ & .01 \\
Qwen3-8B & 20 / 36 & \textbf{.51}$^{*}$ & .46$^{*}$ & .22$^{*}$ & .40$^{*}$ & \textbf{\boldmath$-$.30}$^{*}$ & .07$^{*}$ & .02 \\
\bottomrule
\end{tabular*}
\caption{\label{tab:swaps}
  Swaps: change in expected acceptability when items in the bottom quarter of a factor take the
  position along its direction of items in the top quarter, minus the change for the reverse swap,
  at the layer of Table~\ref{tab:steering}. Bold: the largest positive shift, and overreach where it
  lowers the rating. $^{*}$\,more than two standard errors from zero.}
\end{table*}

Table~\ref{tab:swaps} gives the result of swapping positions along each factor's direction for every factor and model, at the layer of Table~\ref{tab:steering}. Factuality gives the largest shift in five of the six models, up to 1.79 points; in Gemma it is not distinguishable from zero. Swaps along prevalence and cue validity raise the rating in every model, swaps along characteristicness in five, and swaps along overreach lower it in five. Swaps along strikingness shift the rating only slightly, by at most .07 points, although consistently in the same direction, and swaps along the random direction leave it unchanged.

\section{Where people and models depart from the rule}
\label{app:divergence}

For each of the 64 feature-describing items, we compared people's and each model's acceptability rating with what \citeauthor{ryu-etal-2022-flexible}'s rule predicts from their own prevalence and cue-validity ratings. The items the rule gets wrong are mostly the same for people and models, and the models' ratings of the other factors suggest why. Both reject human-related generalizations, such as \textit{Engineers are male} and \textit{Teachers are female}, more than the rule predicts. \citet{ryu-etal-2022-flexible} suggested that people may know more counterexamples about human beings and avoid generalizations that create stereotypes, and the models show this tendency in a sharper form: they agree with people that most engineers are male, yet judge the statement inaccurate and overreaching, treating a common stereotype as a false claim about the group; on average, all twelve reject Ryu et al.'s generalizations about human beings more than the rule predicts. In the other direction, people and models both accept generics such as \textit{Car crashes kill people} and \textit{Ducks lay eggs} more than the rule predicts, and all twelve models rate these as fully accurate, consistent with their judgments following truth rather than frequency. The models are stricter than people with generics whose counterinstances have an alternative property, which \citet{leslie-2008-generics} argues are judged false because they take on an almost universal reading: people are nearly neutral about \textit{Ducks are female} and \textit{Lions are male}, while almost every model rejects both and rates them as inaccurate, as Leslie's account predicts. The smaller models diverge further on generics about harm: they reject \textit{Plastic bags suffocate small children} and \textit{Rip tides drown swimmers}, which people and most of the larger models accept, and five of the six reject \textit{Knives cut people} although all six rate it as accurate. Factuality cannot explain this last rejection; it may reflect safety tuning that makes smaller models reluctant to endorse unqualified statements about harm to people, although they accept others, such as \textit{Car crashes kill people}, so the boundary is not clear-cut. Finally, almost every model accepts \textit{Tables are 10 feet long}, although half of them rate it as inaccurate; here the models may judge whether the sentence is well formed rather than whether it is true, a reading their explanations sometimes adopt.

\end{document}